%% file: neurips_2023.tex
\documentclass{article}

\PassOptionsToPackage{numbers}{natbib}

\usepackage[preprint]{neurips_2023}

\usepackage[utf8]{inputenc} 
\usepackage[T1]{fontenc}    
\usepackage{url}            
\usepackage{booktabs}       
\usepackage{amsfonts}       
\usepackage{nicefrac}       
\usepackage{microtype}      
\usepackage{xcolor}         
\usepackage{subcaption}
\usepackage{graphicx}
\usepackage{lineno}
\usepackage{amsmath, amsthm, bbm, mathtools, commath}
\usepackage{datetime}
\usepackage{colortbl}
\usepackage{arydshln} %

\usepackage{algorithm}
\usepackage{algorithmicx}

\usepackage{algpseudocode}  

\usepackage{indentfirst}
\usepackage{CJK}

\newcommand{\bestcell}{\cellcolor{blue!25}}

\title{Discovering Invariant Neighborhood Patterns for Non-Homophilous Graphs}

\author{Jinluan Yang $^1$, Ruihao Zhang$^1$, \textbf{Zhengyu Chen}$^1$,\\ \textbf{Teng Xiao}$^2$, \textbf{YueyangWang}, \textbf{FeiWu}$^1$, \textbf{Kun Kuang}$^1$
\\
$^1$ Zhejiang University;
$^2$ The Pennsylvania State University;
\\
\texttt{yangjinluan@zju.edu.cn}}

\graphicspath{{photo/}}

\begin{document}
\maketitle

\begin{abstract}
This paper studies the problem of distribution shifts on non-homophilous graphs.
Mosting existing graph neural network methods rely on the homophilous assumption that nodes from the same class are more likely to be linked.
However, such assumptions of homophily do not always hold in real-world graphs, which leads to more complex distribution shifts unaccounted for in previous methods.
The distribution shifts of neighborhood patterns are much more diverse on non-homophilous graphs. 
We propose a novel Invariant Neighborhood Pattern Learning (INPL) to alleviate the distribution shifts problem on non-homophilous graphs.
Specifically, we propose the Adaptive Neighborhood Propagation (ANP) module to capture the adaptive neighborhood information, which could alleviate the neighborhood pattern distribution shifts problem on non-homophilous graphs. We propose Invariant Non-Homophilous Graph Learning (INHGL) module to constrain the ANP and learn invariant graph representation on non-homophilous graphs. Extensive experimental results on real-world non-homophilous graphs show that INPL could achieve state-of-the-art performance for learning on large non-homophilous graphs.
\end{abstract}

\section{Introduction}

Graph Neural Networks (GNNs) have shown promising results in various graph-based applications.
These approaches are based on the strong homogeneity assumption that nodes with similar properties are more likely to be linked together.
Such an assumption of homophily is not always true for heterophilic and non-homophilous graphs. Many real-world applications are non-homophilous graphs, such as online transaction networks \cite{pandit2007fast}, dating networks \cite{zhu2021graph}, molecular networks \cite{zhu2020beyond}, where nodes from different classes tend to make connections due to {opposites attract}.
Recently, various GNNs have been proposed to deal with non-homophilous graphs with different methods \cite{ pei2020geom,abu2019mixhop,zhu2020beyond,chen2020simple,chien2020adaptive,lim2021large, wang2017community}, and these methods heavily rely on the I.I.D assumption that the training and testing data are independently drawn from an identical distribution.
However, these methods are prone to unsatisfactory results when biases occur due to distribution shifts, limiting their effectiveness.

\begin{figure}
\centering 
\includegraphics[width=0.7\linewidth]{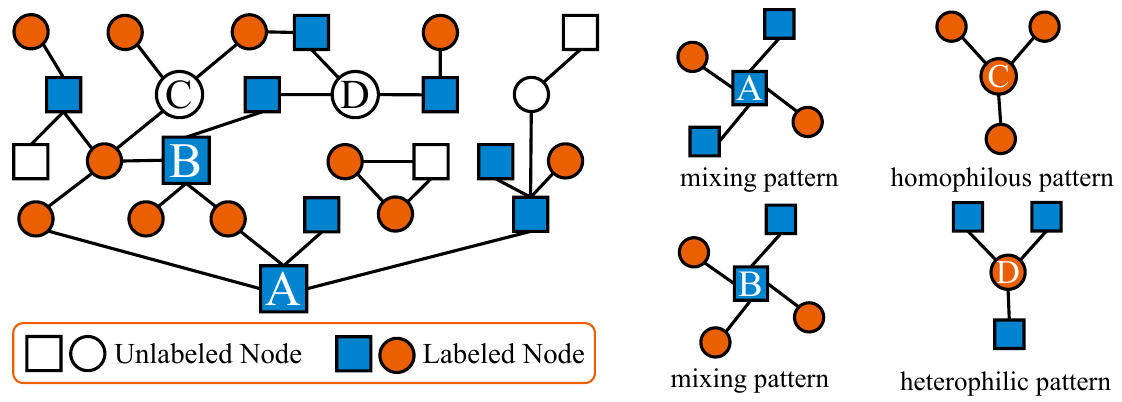}  
\caption{Schematic diagram of the distribution shifts problem
on non-homophilous graphs. The shape denotes the label of each node. 
The shape \emph{circle} is labeled as ``deep learning", and  \emph{rectangle} is labeled as ``system design".
The neighborhood pattern of training nodes A and B are non-homophilous patterns, where parts of nodes are in the same class and the others are not, which dominate the training of GNNs.  
However, the neighborhood pattern of node C is homophilous and the neighborhood pattern of node D is heterophilic, leading to the distribution shifts between training and testing.}    
\label{motivation}       
\end{figure}

To overcome such bias issues caused by distribution shifts, recent works attempt to learn invariant graph representation for GNNs \cite{chen2022ba,wu2022handling,zhao2021graphsmote}. 
These methods tackle bias problems on homophilous graphs, where the bias is caused by degree or class distribution shifts.
Such methods assume that neighboring nodes have similar characteristics.
However, homophily assumptions do not always hold in real-world graphs. 
These methods could not solve bias problems on non-homophilous graphs,
{since the distribution shifts of neighborhood patterns on non-homophilous graphs are much more diverse.}
As shown in Figure \ref{motivation}, the neighborhood pattern of testing node C is homophilous, where the class of node C is the same as the classes of all neighborhoods. In contrast, the neighborhood pattern of testing node D is heterophilic, where all neighborhoods have different classes.
For training nodes, the neighborhood pattern of nodes A and B are \emph{non-homophilous (mixing)}, where parts of nodes are in the same class, and the others are not.
On non-homophilous graphs, a node usually connects with others due to the complex interaction of different latent factors and therefore possesses various neighborhood pattern distributions wherein certain parts of the neighborhood are homophilous while others are heterophilic, which leads to the \emph{neighborhood pattern distribution shifts} problem.

\begin{figure*}
\centering
\begin{subfigure}{0.24\linewidth}
     \includegraphics[scale=0.17]{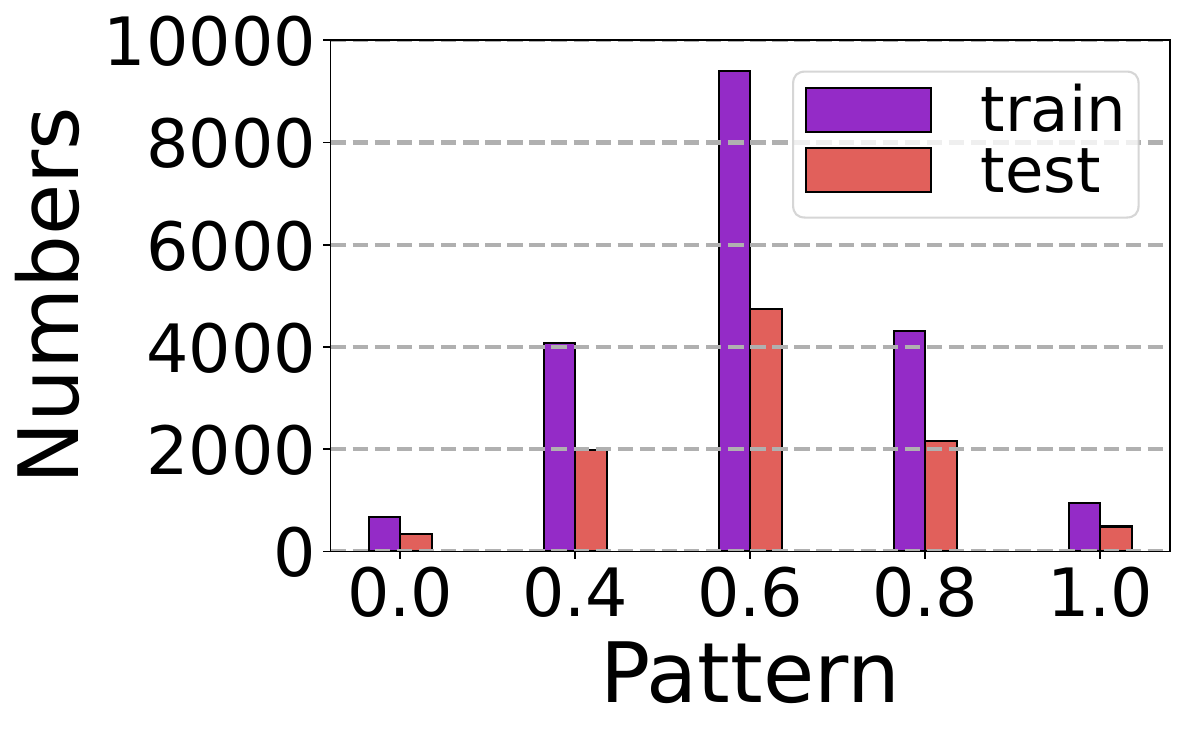}  
     \caption{\small Pattern distribution}
\end{subfigure}
\begin{subfigure}{0.24\linewidth}
     \includegraphics[scale=0.17]{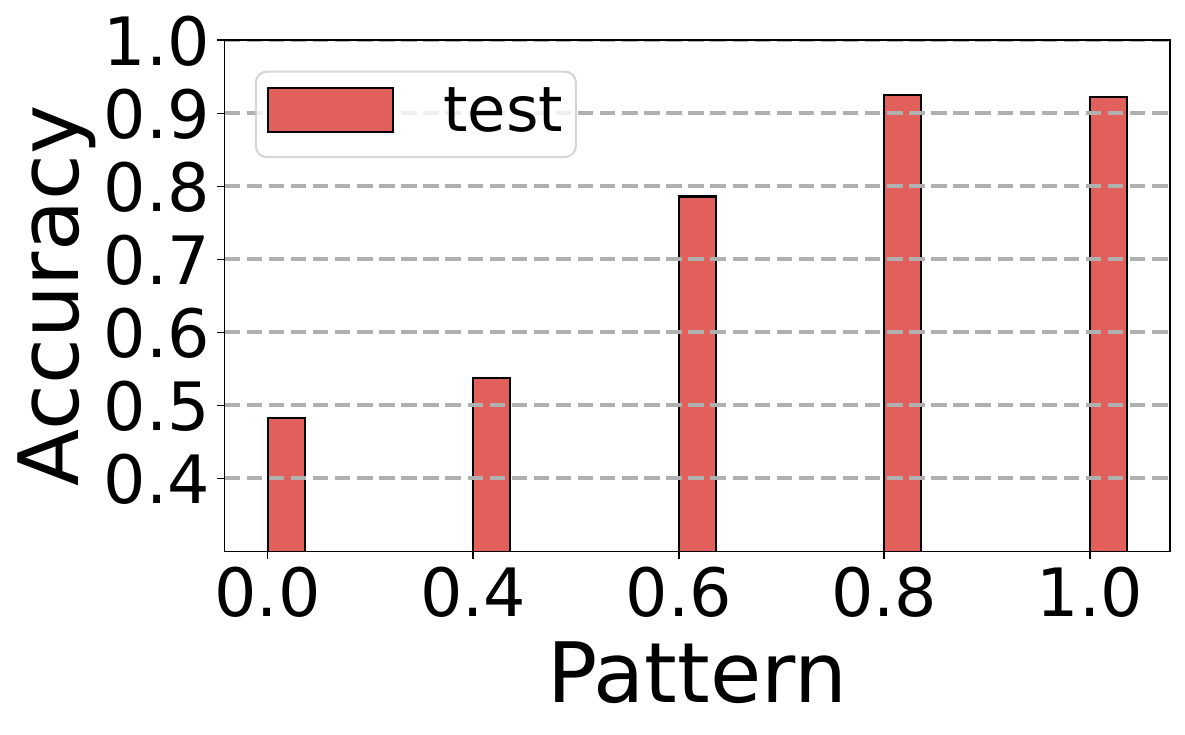}  
     \caption{\small Pattern performance}
\end{subfigure}
\begin{subfigure}{0.24\linewidth}
     \includegraphics[scale=0.17]{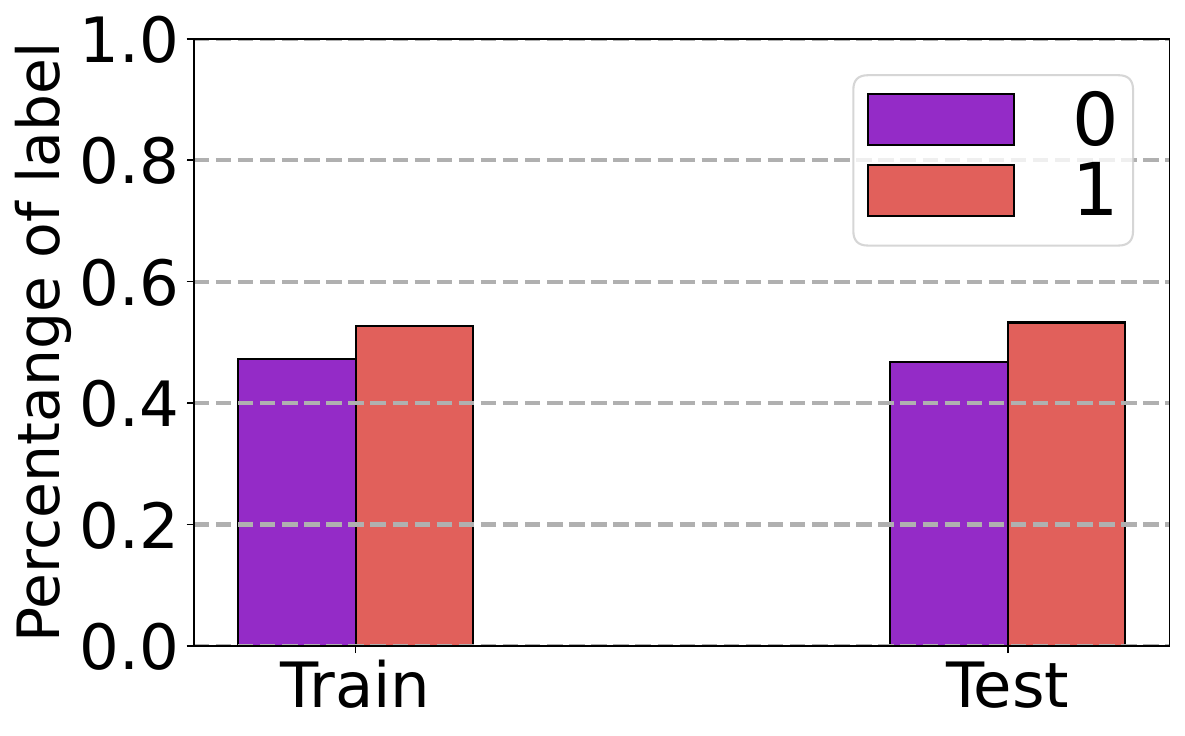}  
     \caption{Label distribution}
\end{subfigure}
\begin{subfigure}{0.24\linewidth}
     \includegraphics[scale=0.17]{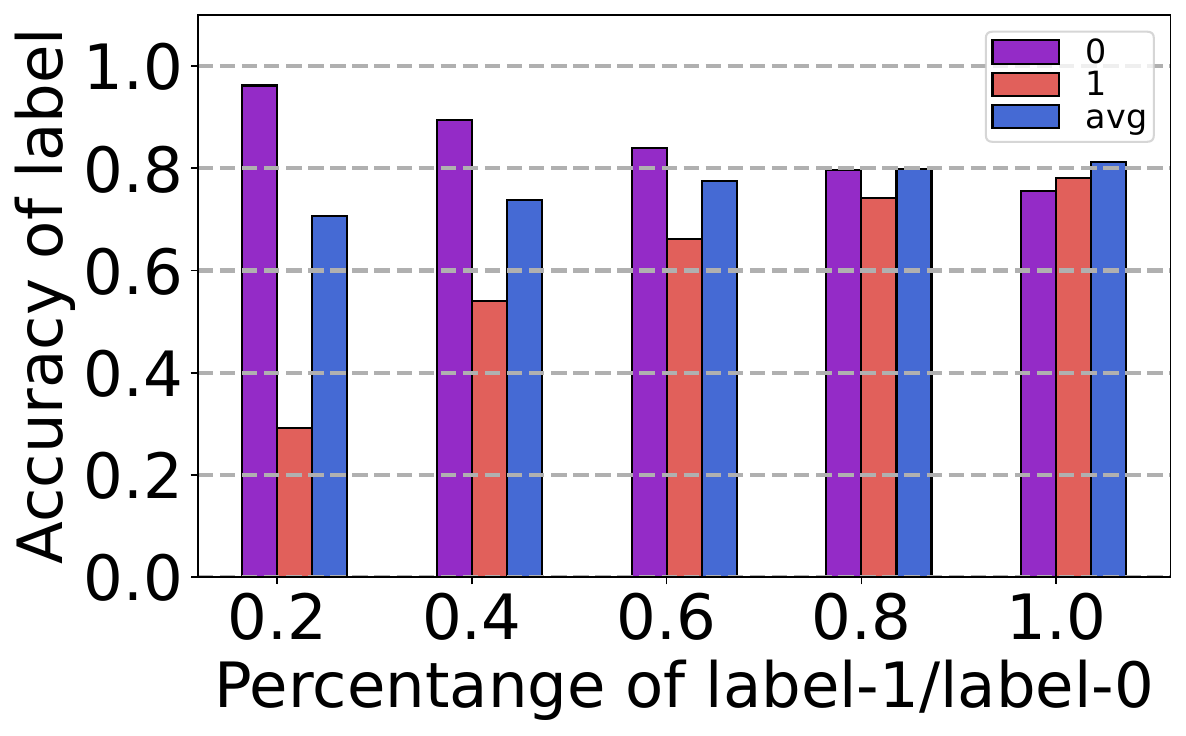}  
     \caption{Label performance}
\end{subfigure}
  
\caption{Empirical investigation of distribution shifts in Penn94. (a) shows the pattern distributions, and the performance of node classification with different neighborhoods is shown in (b). (c) shows the label distribution, the performance of node classification with different labels is shown in (d).}
    \label{fig: neighborhood distribution}
\end{figure*}

To further verify if the \emph{neighborhood pattern distribution shifts} between training and testing can impair the performance of GNNs on non-homophilous graphs, we conduct an empirical investigation.
Figure \ref{fig: neighborhood distribution} (a) shows the neighborhood distribution of Penn94 dataset, {where pattern 0 is the heterophilic node (0\% neighborhood nodes are in the same class)} and pattern 1 is the homophilous node (100\% neighborhood nodes are in the same class), others patterns are non-homophilous nodes.
Most nodes are in mixing patterns rather than homophilous or heterophilic patterns, which leads to diverse neighborhood distribution.
Results on Figure \ref{fig: neighborhood distribution} (b) show that the performance of {GCN} on nodes with different patterns varies considerably across the graph. 
Moreover, 
the test distribution remains unknown during the training of GNN, heightening uncertainty regarding generalization capabilities.


To address such \emph{unknown neighborhood distribution shifts} problem, we are faced with two main challenges. The first challenge is how to alleviate the neighborhood pattern distribution shifts problem. The neighborhood distributions typically have diverse patterns, where nodes connect with other nodes on non-homophilous graphs due to the intricate interaction of various latent factors, giving rise to a variety of neighborhood pattern distributions. The second challenge is how to alleviate the distribution shifts in unknown test environments. 
Figure \ref{fig: neighborhood distribution} (c) shows the label distribution shifts also existing in Penn94, and leads to poor performance of GNN in Figure  \ref{fig: neighborhood distribution} (d). 
Previous works \cite{chen2022ba,wu2022handling,zhao2021graphsmote,zhou2019graph,tang2020investigating} ignore these unknown biases, especially neighborhood distribution shifts on non-homophilous graphs.

This paper presents Invariant Neighborhood Pattern Learning (INPL) framework that aims to alleviate the distribution shifts on non-homophilous graphs. Specifically, we propose the Adaptive Neighborhood Propagation (ANP) module to capture the adaptive neighborhood information and Invariant Non-Homophilous Graph Learning (INHGL) module to constrain the ANP and learn invariant graph representation on non-homophilous graphs.
Our contributions can be summarized as follows:  (1) We study a novel bias problem caused by neighborhood pattern distribution shifts on non-homophilous graphs. (2) We design a scalable framework Invariant Neighborhood Pattern Learning (INPL) to alleviate unknown distribution shifts on non-homophilous graphs, which learns invariant representation for each node and makes invariant predictions on various unknown test environments.
(3) We conduct experiments on eleven real-world non-homophilous graphs, and the results show that INPL could achieve state-of-the-art performance for learning on large non-homophilous graphs.

\section{Related work}
\label{Related work}
\textbf{Generalization on non-homophilous Graphs.} Recently some non-homophilous methods have been proposed to make GNNs generalize to non-homophilous graphs. Generally, these non-homophilous GNNs enhance the performance of GNNs through the following three main designs: (1) Using High-Order Neighborhoods. (2) Separating ego- and neighbor-embedding. (3) Combining Inter-layer representation.
Geom-GCN \cite{pei2020geom} aggregates immediate neighborhoods and distant nodes that have a certain similarity with the target node in a continuous space. 
MixHop \cite{abu2019mixhop} Proposes a mixed feature model which aggregates messages from multi-hop neighbors by mixing powers of the adjacency matrix.
GPR-GNN \cite{chien2020adaptive} adaptively learns the GPN weights to extract node features and topological information. 
H2GCN \cite{zhu2020beyond} applies three useful designs—ego- and neighbor-embedding separation, higher-order neighborhoods and a combination of intermediate representation that boost learning from the graph structure under low-homophily settings.
GCNII \cite{chen2020simple} applies initial residuals and constant mapping to relieve the problem of over-smoothing, which empirically performs better in non-homophilous settings.
LINKX \cite{lim2021large} proposes a simple method that combines two simple baselines MLP and LINK, which achieves state-of-the-art performance 
while overcoming the scalable issues.
However, these works ignore the distribution shifts problem on non-homophilous graphs. Different from these works, {we focus on neighborhood distribution shifts on non-homophilous graphs.}


\textbf{Debiased Graph Neural Network.} Debiased Graph Neural Networks aim to address the bias issue on graphs \cite{chen2024learning,yang2025leveraging,lv2025grasp,hu2024let}.
To overcome degree-related bias, 
SL-DSGCN \cite{tang2020investigating} mitigates the degree-related bias of GCNs by capturing both discrepancies and similarities of nodes with different degrees.
ImGAGN \cite{qu2021imgagn} is proposed to address the class-related bias, ImGAGN generates a set of synthetic minority nodes to balance the class distribution.
Different from these works on a single bias, 
BA-GNN \cite{chen2022ba} proposes a novel Bias-Aware Graph Neural Network (BA-GNN) framework by learning node representation that is invariant across different biases and distributions for invariant prediction.
EERM \cite{wu2022handling} 
facilitates graph neural networks to leverage invariance principles for prediction, EERM 
resorts to multiple context explorers that are adversarially trained to maximize the variance of risks from multiple virtual environments, which enables the model to extrapolate from a single observed environment which is the common case for node-level prediction. However, these works heavily rely on the assumption of homophily, which may be prohibitively failed on non-homophilous graphs.
Our method differs from the above methods and aims to learn invariant graph representation for non-homophilous graphs.


\section{Discovering Invariant Neighborhood Patterns}
\label{FrameWork}

We propose a novel Invariant Neighborhood Pattern Learning (INPL) framework that aims to alleviate the distribution shifts on non-homophilous graphs.
1) To alleviate the neighborhood pattern distribution shifts problem on non-homophilous graphs, we propose Adaptive Neighborhood Propagation, where the Invariant Propagation layer is proposed to combine both the high-order and low-order neighborhood information. And adaptive propagation is proposed to capture the adaptive neighborhood information.
2) To alleviate the distribution shifts in unknown test environments, we propose Invariant Non-Homophilous Graph Learning to constrain the Adaptive Neighborhood Propagation, which learns invariant graph representation on non-homophilous graphs. Specifically,
we design the environment clustering module to learn multiple graph partitions, and the invariant graph learning module learns the invariant graph representation based on multiple graph partitions. 
 

\paragraph{Problem formulation.} Let $G = (V, E, X)$ be a graph, where $V$ is the set of nodes, $E$ is the set of edges, $X$ $\in$ $R^{N \times d}$ is a feature matrix, each row in $X$ indicates a d-dimensional vector of a node.  $A$ $\in$ $\{0,1\}^{N \times N}$ is the adjacency matrix, where $A_{uv} = 1$ if there exists an edge between node $v_i$ and $v_j$, otherwise $A_{ij} = 0$ . For semi-supervised node classification tasks, only part of nodes have known labels $Y^o = \{y_1,y_2,...y_n\}$, where $y_j \in \{0,1,...,c-1\}$ denotes the label of node $v_j$, $c$ is the number of classes. 
Similar to \cite{chen2022ba}, we also define a graph environment to be a joint distribution $P_{XAY}$ on $X \times A \times Y$ and let $\mathcal{E}$ denote the set of all environments. In each environment $e \in \mathcal{E}$, we have a graph dataset $G^e = (X^e, A^e, Y^e)$, where $X^e \in \mathcal{X}$ are node features, $A^e \in \mathcal{A}$ is the adjacency matrix and $Y^e \in \mathcal{Y}$ is the response variable. The joint distribution $P_{XAY}^e$ of $X^e$, $A^e$ and $Y^e$ can vary across environments: $P_{XAY}^e \ne P_{XAY}^{e^\prime}$ for $e,e^\prime \in \mathcal{E}$ and $e \ne e^\prime$. In this paper, our goal is to learn a graph neural network, which can make invariant prediction across unknown environments on non-homophilous graphs, we define the node classification problem on non-homophilous graphs as:

Given a training graph $\mathcal{G}_{train} = \{A_{train}, X_{train}, Y_{train}\}$, the task is to learn a GNN $g_\theta(\cdot)$   with parameter $\theta$ to precisely predict the label of nodes on different test environments $\{\mathcal{G}_{test}^1,\mathcal{G}_{test}^2,\cdot\cdot\cdot,\mathcal{G}_{test}^e\}$, where $\mathcal{G}_{test} = \{A_{test}, X_{test}, Y_{test}\}$.




\begin{figure*}
\centering
\includegraphics[width=14cm]{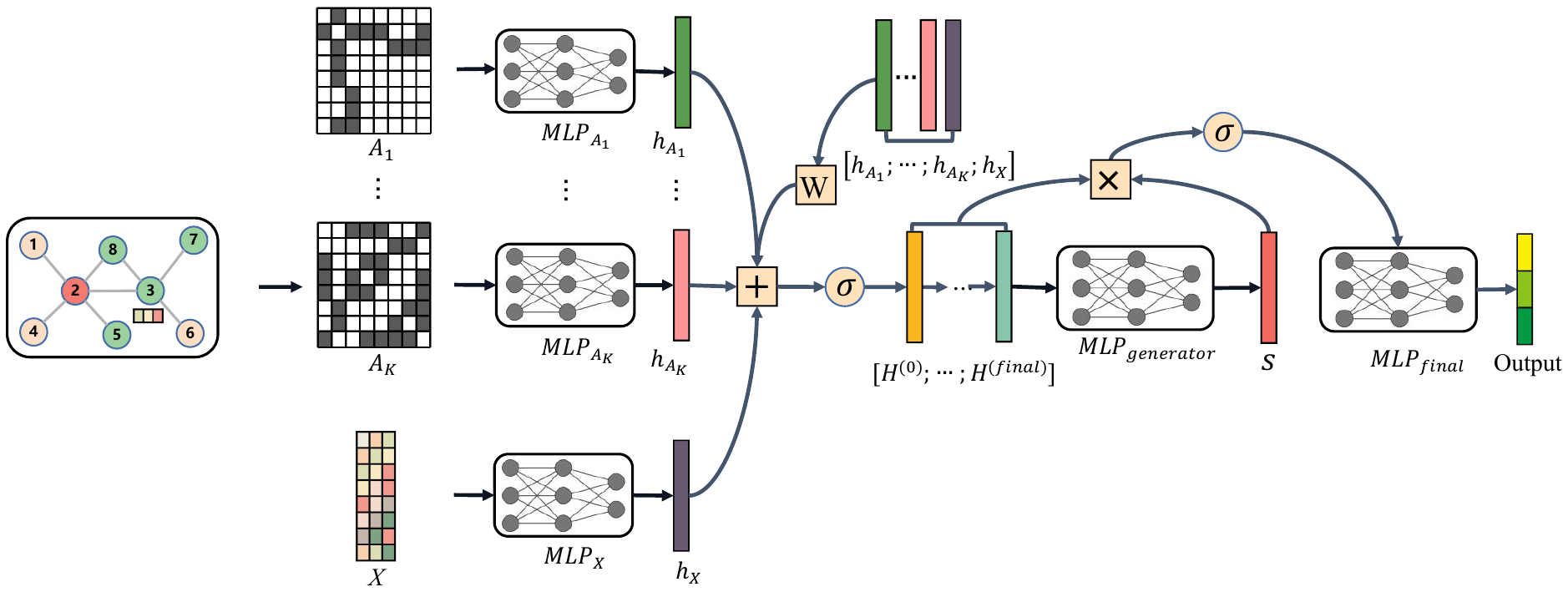}  
\caption{The module of Adaptive Neighborhood Propagation (ANP), the invariant propagation layer is proposed to combine both the high-order and low-order neighborhood information. And adaptive propagation is proposed to capture the adaptive neighborhood information, $s$ is the adaptive propagation step learned by the generative network $q_{\phi}$.
}    
\label{fig:method}       
\end{figure*}

\noindent

\subsection{Adaptive Neighborhood Propagation}

\paragraph{Invariant Propagation Layer.}
To combine both the high-order and low-order neighborhood information, 
we design the invariant propagation layer as:
\begin{align}
    H^{(l+1)} = \sigma \left(\left( \left( 1-\alpha_{l} \right)  H^{(l)}  + \alpha_{l}H^{(0)}\right) \left( \left( 1-\beta_l\right)I_n +\beta_{l}W^{(l)}\right)\right)
\end{align}
where $\alpha_l$ and $\beta_l$ are two hyperparameters, $H^{(0)} = \sigma(W[h_{A_1}; h_{A_2};\cdot\cdot\cdot;h_{A_K};h_X] + h_{A_1} + h_{A_2} + \cdot\cdot\cdot + h_{A_K} + h_X)$, $h_X$ are embedding obtained by MLPs processing of node features, $A_K$ is the $K$th-order adjacency matrix of graphs, $h_{A_K}$ is the embedding information obtained by processing the adjacency matrix $A_K$ by MLPs. $I_n$ is an identity mapping and $W^{(l)}$ is $l$-th weight matrix. 


\emph{High-order neighborhoods}. On non-homophilous graphs, nodes with semantic similarity to the target node are usually higher-order neighborhood nodes, so here we leverage high-order neighborhood information, which is a common strategy for non-homophilous methods \cite{abu2019mixhop, zhu2020beyond}.

\emph{Initial residual}. The initial residual connection ensures that the ﬁnal representation of each node retains at least a fraction of $\alpha_l$ 
 from the input layer even if we stack many layers \cite{chen2020simple}.

\emph{Identity mapping}. 
Identity mapping plays a significant role in enhancing performance in semi-supervised tasks \cite{he2016deep,chen2020simple}.
Application of identity mappings to some classifiers results in effective constraining of the weights $W^{(\ell)}$, which could
avoid over-ﬁtting and focus on global minimum \cite{hardt2016identity}.

{Similar to LINKX, Invariant Propagation Layer (IPL) is a scalable graph learning method. Since IPL combines both low-order and high-order neighborhoods by using low-order and high-order propagation matrices, which can be precomputed before training, then uses mixing weights to generate the embeddings of each layer after the 0-th layer, thus it can scale to bigger datasets like LINKX, extensive experiments below on large datasets evaluate the scalability of IPL. However, how to learn representation with proper neighborhood information is still a challenge. We propose adaptive propagation to capture the adaptive neighborhood information.
}

\begin{figure*}
\centering 
\includegraphics[width=14cm]{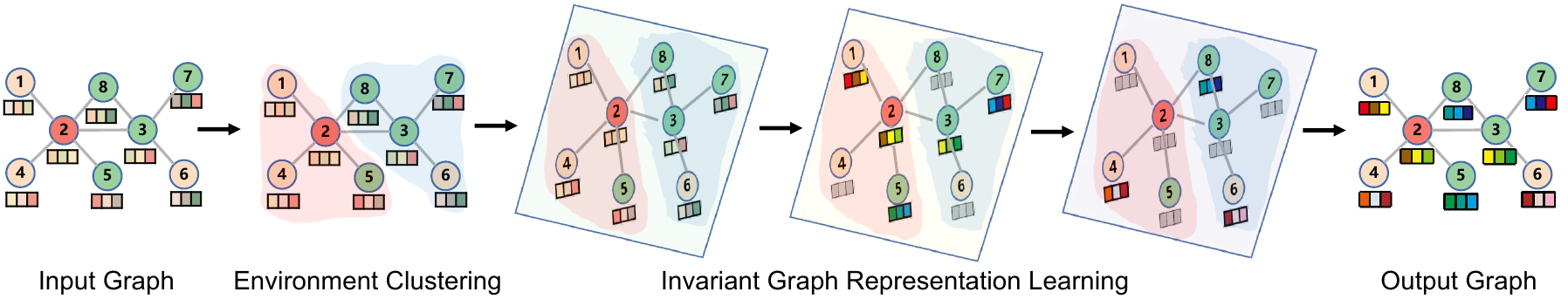}  
\caption{The illustration example of Invariant Non-Homophilous Graph Learning (INHGL). INHGL constrains the ANP to alleviate the distribution shifts in unknown environments. 
}    
\label{fig:framework}       
\end{figure*}

\paragraph{Adaptive Propagation.}
 In practice, there is no way to directly access the optimal propagation of neighborhood information. To capture the adaptive neighborhood information, we learn an adaptive propagation step $s$ by generative network $q_{\phi}$.
Specifically, we infer the optimal propagation distribution $q_{\phi}(s_n| \mathbf{X},\mathbf{A})$  and learn GNN weights $\theta$ jointly with the loss function.
 However, the optimal propagation steps  $s$ are discrete and non-differentiable which makes direct optimization difficult. Therefore, we adopt Gumbel-Softmax Sampling~ \cite{jang2016categorical,maddison2016concrete}, which is a simple yet effective way to
substitutes the original non-differentiable sample from a discrete distribution with a differentiable sample from a corresponding Gumbel-Softmax distribution. 
Thus, we have the loss function as:
\begin{linenomath}
\small
\begin{align}
 \mathcal{L}({\phi}) =-\log p_{{\theta}}(\mathbf{y} | GNN (\mathbf{X},\mathbf{A},\hat{s}))+\text{KL}(q_{\phi}(s_n|\mathbf{X},\mathbf{A}) ||p(s_n)), \label{Eq:ELBO2}
\end{align}
\end{linenomath}
where $\hat{s}$ is drawn  from a categorical distribution with the discrete variational  distribution $q_{{\phi}}(s_n|\mathbf{X},\mathbf{A})$ parameterized by $\phi$:
\begin{linenomath}
\small
\begin{align}
\hat{s}_{k}=\frac{\exp (( \log ( q_{{\phi}}(s_n|\mathbf{X},\mathbf{A})[a_k])+g_{k}) / \gamma_g )}{\sum_{k'=0}^{K} \exp ( ( \log(q_{{\phi}}(s_n|\mathbf{X},\mathbf{A})[a_{k'}])+g_{k'}) / \gamma_g )},
\end{align}
\end{linenomath}
where $\left\{g_{k'}\right\}_{k'=0}^{K}$ are i.i.d. samples drawn from the Gumbel (0, 1) distribution, $\gamma_g$ is the softmax
temperature, $\hat{s}_{k}$ is the $k$-th value of sample $\hat{s}$ and  $q_{{\phi}}(s_n|\mathbf{X},\mathbf{A})[a_k]$  indicates the $a_{k}$-th index of $q_{{\phi}}(s_n|\mathbf{X},\mathbf{A})$, i.e., the logit corresponding the  $(a_{k}-1)$-th layer. Clearly, when $\tau>0$, the Gumbel-Softmax distribution is smooth so $\phi$ can be  optimized by standard back-propagation. The  KL term in Eq.~(\ref{Eq:ELBO2}) is respect to two categorical distributions, thus it has a closed form.

\subsection{Invariant non-homophilous Graph Learning}

The proposed Adaptive Neighborhood Propagation (ANP) module could alleviate the neighborhood pattern distribution shifts problem on non-homophilous graphs by capturing the adaptive neighborhood information.
However, as illustrated in Figure \ref{motivation}, the testing environments are always unknown and unpredictable, where the class distributions and neighborhood pattern distribution are various. Such unknown distribution shifts would render GNNs over-optimized on the labeled training samples, which hampers their capacity for robustness and leads to a poor generalization of GNNs. 

Inspired by previous invariant learning methods \cite{arjovsky2019invariant,peters2015causal}, we propose Invariant Non-Homophilous Graph Learning to overcome such unknown distribution shifts, which learns invariant graph representation on non-homophilous graphs. To surmount these issues, two modules form part of our solution strategy. Specifically,
we design the environment clustering module to learn multiple graph partitions, and the invariant graph learning module learns the invariant representation based on these partitions.

In order to learn invariant graph representation across different environments, we have a commonly used assumption in previous invariant learning methods \cite{arjovsky2019invariant,peters2015causal}:

$Assumption$: we assumes that there exists a graph representation $\Phi(X,A)$, for all environments $e, e^{\prime} \in \mathcal{E}$
, we have $P[Y^e|\Phi(X^e, A^e))]=P[Y^{e^\prime}|\Phi(X^{e^\prime}, A^{e^\prime}))]$.

This assumption shows that we could learn invariant graph representation with a proper graph network $\Phi$ across different environments. 
However, the majority of available graphs do not have environmental labels. Thus, we generate multiple graph partitions as different environments by the environment clustering module, and learn invariant graph representation in these different environments.

\paragraph{Environment Clustering.}
 The environment clustering module takes a single graph as input and outputs multiple graph environments. The goal of the environment clustering module is to increase the similarity of nodes within the same environment while decreasing the bias between different environments.
 Thus, the nodes should be clustered by the variant relation between the node and the target label. And the variant information $\Psi(X, A)$ is updated with ANP as $\Psi(X, A) = ANP(X, A)$. We also random select $K$ cluster centroids $\mu_{1},\mu_{2},...,\mu_{k} \in \mathcal{R}^{n}$, thus there exists $K$ cluster $G^{1},G^{2},...,G^{K} \in \mathcal{E}_{tr}$ at the beginning of optimization.
Here we choose K-means as the method for environmental clustering, which is one of the representative methods of unsupervised clustering. 
The objective function of the environment clustering module is to minimize the sum of the distances between all nodes and their associated cluster centroids $ \mathcal{L}= \min\frac{1}{N}\sum_{j=1}^{K}\sum_{i=1}^{N} \left \| {\Psi(X_i,A_i)} - \mu_j \right \|^2$.

For environments $\mathcal{E}_{tr}$, each node $i$ is assigned to environment $G^{e} \in \mathcal{E}_{tr}$ with the closest cluster center by $ G^{i} := argmin_{e}\left\|{\Psi(X_i,A_i)} -\mu_e \right\|^2, e\in 1,2,...,K$.
For every environment $G^{e} \in \mathcal{E}_{tr}$, the value of this cluster center $\mu_e$ is updated as $\mu_e := \frac{\sum_{i=1}^m l\{G^{i}=e\}{\Psi(X_i,A_i)}}{\sum_{i=1}^ml\{G^{i}=e\}}$.

\paragraph{Invariant Graph Representation Learning.}

The invariant graph representation learning takes multiple graphs partitions $G = {G^e}_{e\in supp(\mathcal{E}_{tr})}$ as input, {and learns invariant graph representation. }

To learn invariant graph representation on non-homophilous graphs, we capture adaptive neighborhood information, where we use ANP to make predictions as
$\hat{Y}^e = ANP(X^e,A^e)$. 
To learn invariant graph representation, we use the variance penalty similar to \cite{DBLP:journals/corr/abs-2008-01883}, The objective function of INPL is:
\begin{equation}\label{eq:lossmain}
    \mathcal{L}_p(G;\theta) = E_{\mathcal{E}_{tr}}(\mathcal{L}^e) + \lambda * Var_{\mathcal{E}_{tr}}(\mathcal{L}^e) =  E_{\mathcal{E}_{tr}}(\mathcal{L}\left(\hat{Y}^e ,Y^e\right)) + \lambda * Var_{\mathcal{E}_{tr}}(\mathcal{L}\left(\hat{Y}^e ,Y^e\right))
\end{equation}
With such a penalty, we could constrain ANP to alleviate distribution shifts in unknown environments.

\begin{table*}
\captionsetup{font={small,stretch=1.25}, labelfont={bf}}
 \renewcommand{\arraystretch}{1}
    \caption{Experimental results. The three best results per dataset is highlighted. (M) denotes out of memory.}
    \label{tab:results}
    \centering
        \resizebox{0.99\linewidth}{!}{
    \begin{tabular}{cccccccccccc}
        \toprule[1.5pt]
&snap-patents &pokec &genius& twitch-gamers &  Penn94 &   film & squirrel & chameleon & cornell & texas & wisconsin \\
        \#Nodes &2,923,922 &1,632,803 &421,961& 168,114 & 41,554 &7,600 & 5,201&2,277 &251 &183 & 183 \\
        
   \toprule[1.0pt]
MLP &$31.34\pm{0.05}$ &$62.37\pm{0.02}$ &$86.68 \pm{0.09}$ & $60.92\pm{0.07}$  & $73.61\pm{0.40}$ &$34.50 \pm{1.77}$  & $31.10\pm{0.62}$ & $41.67 \pm{5.92}$  & $67.03 \pm{6.16} $& $70.81 \pm{4.44} $ & $71.77 \pm{5.30}$\\    
     LINK &\bestcell$60.39\pm{0.07}$ &$80.54\pm{0.03}$ &$73.56 \pm{0.14}$ & $64.85\pm{0.21}$ & $80.79\pm{0.49}$ & $23.82 \pm{0.30}$ & $59.75\pm{0.74}$ &\bestcell $64.21\pm{3.19}$& $44.33\pm{3.63} $ & $51.89\pm{2.96} $ & $54.90\pm{1.39} $  \\    
     \hdashline
     GCN &$45.65\pm{0.04}$ &$75.45\pm{0.17}$ &$87.42\pm{0.37}$& $62.18\pm{0.26}$ & $82.47\pm{0.27}$ & 26.86  & 23.96 & 28.18 & 52.70 & 52.16 & 45.88  \\
     GAT &$45.37\pm{0.44}$ &$71.77\pm{6.18}$ &$55.80\pm{0.87}$& $59.89\pm{4.12}$ & $81.53\pm{0.55}$ & 28.45 & 30.03 & 42.93 & 54.32 & 58.38 & 49.41\\
     GCNJK &$46.88\pm{0.13}$ &$77.00\pm{0.14}$ &$89.30\pm{0.19}$& $63.45\pm{0.22}$ & $ 81.63\pm{0.54} $ & 27.41 & 35.29  & 57.68   & 57.30 & 56.49 & 48.82 \\
     APPNP &$32.19\pm{0.07}$ &$62.58\pm{0.08}$ &$85.36\pm{0.62}$& $60.97\pm{0.10}$ & $74.33\pm{0.38}$  & 32.41  & 34.91 & 54.3 & 73.51 & 65.41 & 69.02 \\
   \hdashline
    SL-DSGCN & - & - & - &  $59.28\pm{0.27}$ &   $79.48\pm{0.81}$ &  $26.17\pm{2.09}$  &  $30.42\pm{1.56}$ & \ $36.40\pm{1.64}$ & $54.08\pm{3.89}$ &  $53.52\pm{6.17}$ &  $45.24\pm{6.04}$ \\ 
  
  ImGAGN & - & - & - &  $60.78\pm{0.36}$ &   $80.65\pm{0.47}$ &  $25.32\pm{1.23}$  &  $31.05\pm{1.54}$ & \ $40.65\pm{1.16}$ & $54.91\pm{6.26}$ &  $54.87\pm{7.12}$ &  $49.14\pm{6.96}$ \\ 
  
 BA-GNN & - & - & - & $62.82\pm{0.29}$ & $83.46\pm{0.57}$ & 
$27.62\pm{1.52}$  & $32.97\pm{1.56}$  & $43.97\pm{1.70}$  
& $55.19\pm{5.11}$  & $54.98\pm{6.85}$ & $49.70\pm{7.67}$\\

     \hdashline
     H\textsubscript{2}GCN-1 &(M) &(M) &(M)& (M) & (M)  &\bestcell $35.86\pm{1.03}$ & $36.42\pm{1.89}$ & $57.11\pm{1.58}$ &\bestcell $82.16\pm{4.80}$ &\bestcell $84.86\pm{6.77}$ &\bestcell  $86.67\pm{4.69}$\\
     
       H\textsubscript{2}GCN-2 &(M) &(M) &(M)& (M) & (M)  & $35.62\pm{1.30}$ & $37.90\pm{2.02}$ & $59.39\pm{1.98}$ & \bestcell$82.16\pm{6.00}$ &\bestcell $82.16\pm{5.28}$ &\bestcell  $85.88\pm{4.22}$\\   
     MixHop &$52.16\pm{09}$ &\bestcell$81.07\pm{0.16}$ &\bestcell$90.58\pm{0.16}$&\bestcell  $65.64\pm{0.27}$ &\bestcell  $83.47\pm{0.71}$ &  $32.22\pm{2.34}$  & $43.80\pm{1.48}$ &  $60.50\pm{2.53}$ &  $73.51\pm{6.34}$ & $77.84\pm{7.73}$ & $75.88\pm{4.90}$    \\
     GPR-GNN &$40.19\pm{0.03}$ &$78.83 \pm{0.05}$ &$90.05 \pm{0.31}$& $61.89\pm{0.29}$ & $81.38 \pm{0.16}$ & $33.12\pm{0.57}$ &  $54.35\pm{0.87}$  & $62.85\pm{2.90}$ & $68.65\pm{9.86}$ & $76.22\pm{10.19}$ & $75.69\pm{6.59}$\\
	 GCNII &$37.88\pm{0.69}$ &$78.94 \pm{0.11}$ &$90.24 \pm{0.09}$& $63.39\pm{0.61}$ & $82.92\pm{0.59}$ & $34.36 \pm{0.77}$  &\bestcell $56.63 \pm{1.17}$ & 62.48 & 76.49 & 77.84 & 81.57 \\
     Geom-GCN-I &(M) &(M) &(M)& (M) & (M) & 29.09 & 33.32 & 60.31 & 56.76 & 57.58 & 58.24 \\
     Geom-GCN-P &(M) &(M) &(M)& (M) & (M) & 31.63 & 38.14 & 60.90 & 60.81 & 67.57 & 64.12 \\
     Geom-GCN-S &(M) &(M) &(M)& (M) & (M) & 30.30 & 36.24 & 59.96 & 55.68 & 59.73 & 56.67 \\
	 LINKX  &\bestcell$61.95 \pm{0.12}$ &\bestcell$82.04 \pm{0.07}$ &\bestcell$90.77 \pm{0.27}$&\bestcell  $66.06\pm{0.19}$ &\bestcell   $84.71\pm{0.52}$ &\bestcell  $36.10\pm{1.55}$  &\bestcell  $61.81\pm{1.80}$ &\bestcell \ $68.42\pm{1.38}$ & $77.84\pm{5.81}$ &  $74.60\pm{8.37}$ &  $75.49\pm{5.72}$ \\
  \toprule[1.5pt]
   \textbf{INPL} &\bestcell $62.49\pm{0.05}$ &\bestcell$83.09\pm{0.05}$ &\bestcell$91.49\pm{0.07}$ &\bestcell   $66.75\pm{0.07}$ &\bestcell  $86.20\pm{0.05}$  &\bestcell  $38.12\pm{0.36}$ &\bestcell  $64.38\pm{0.62}$ &\bestcell  $71.84\pm{1.22}$ &\bestcell  $83.24\pm{1.21}$ & \bestcell $84.86\pm{3.08}$ &\bestcell  $85.88\pm{0.88}$ \\
        \toprule[1.5pt]    
    \end{tabular}}
\end{table*}

\begin{figure*}
\centering
\begin{subfigure}{0.31\linewidth}
     \includegraphics[scale=0.22]{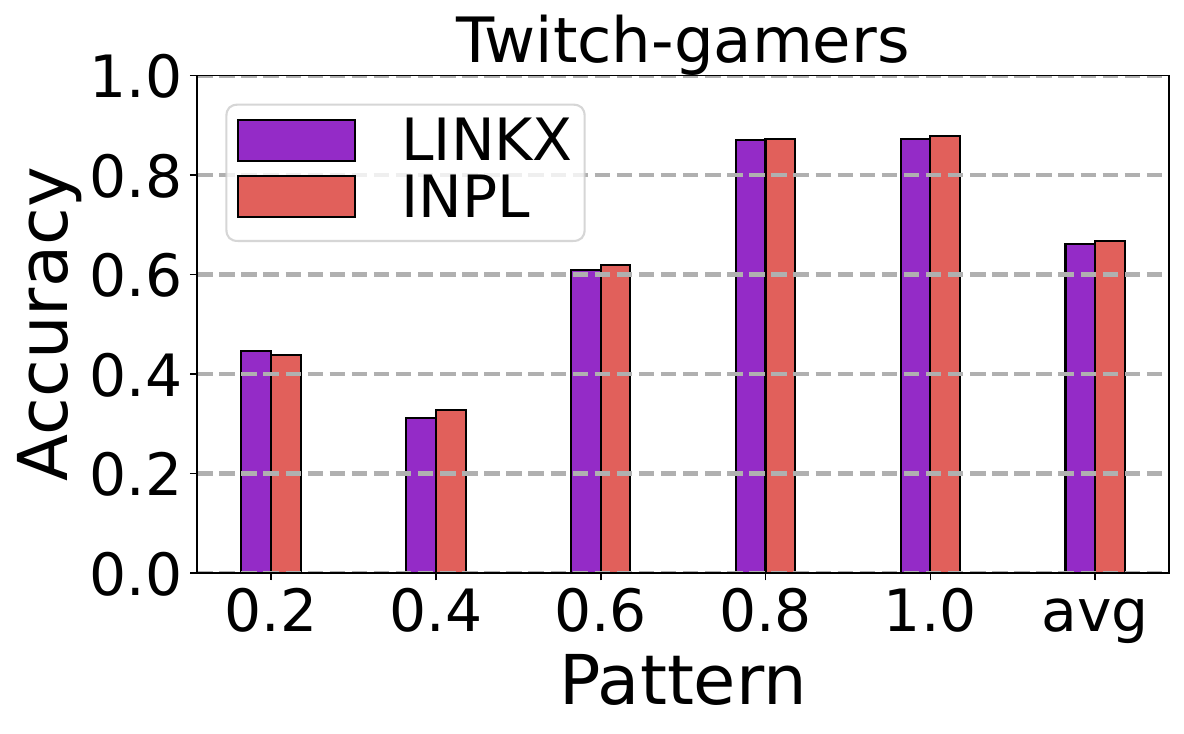}  
\end{subfigure}
\begin{subfigure}{0.31\linewidth}
     \includegraphics[scale=0.22]{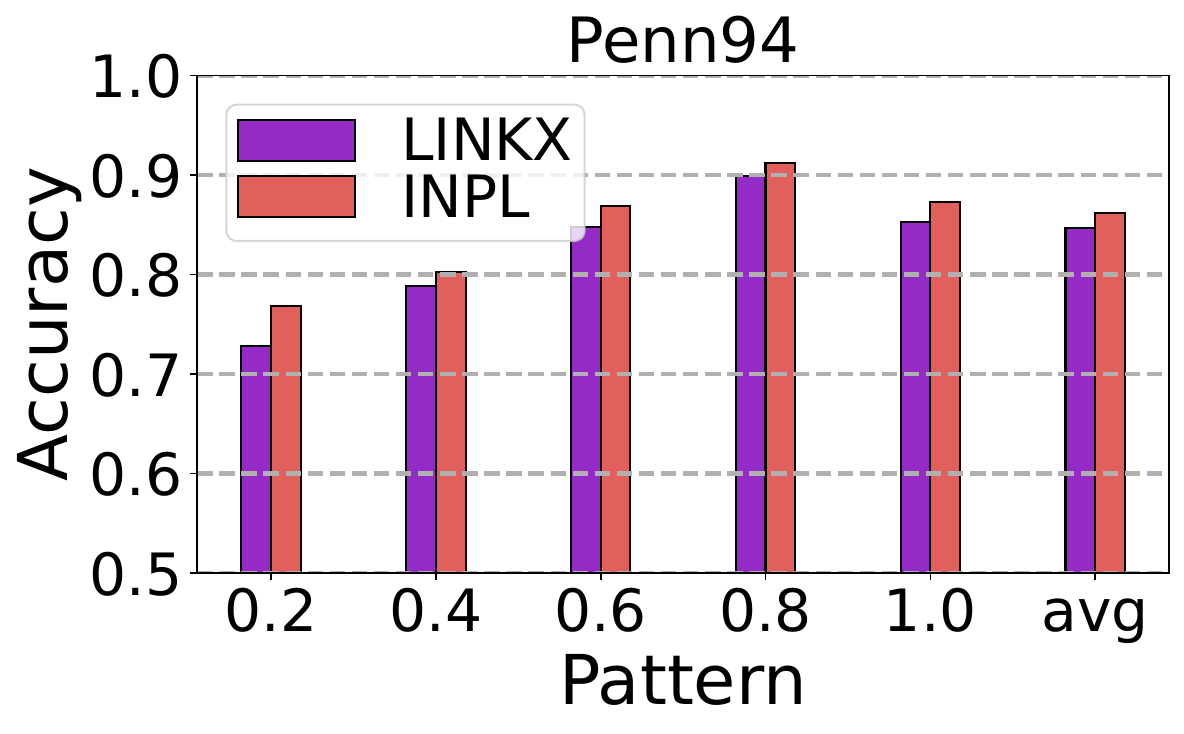}  
\end{subfigure}
\begin{subfigure}{0.31\linewidth}
     \includegraphics[scale=0.22]{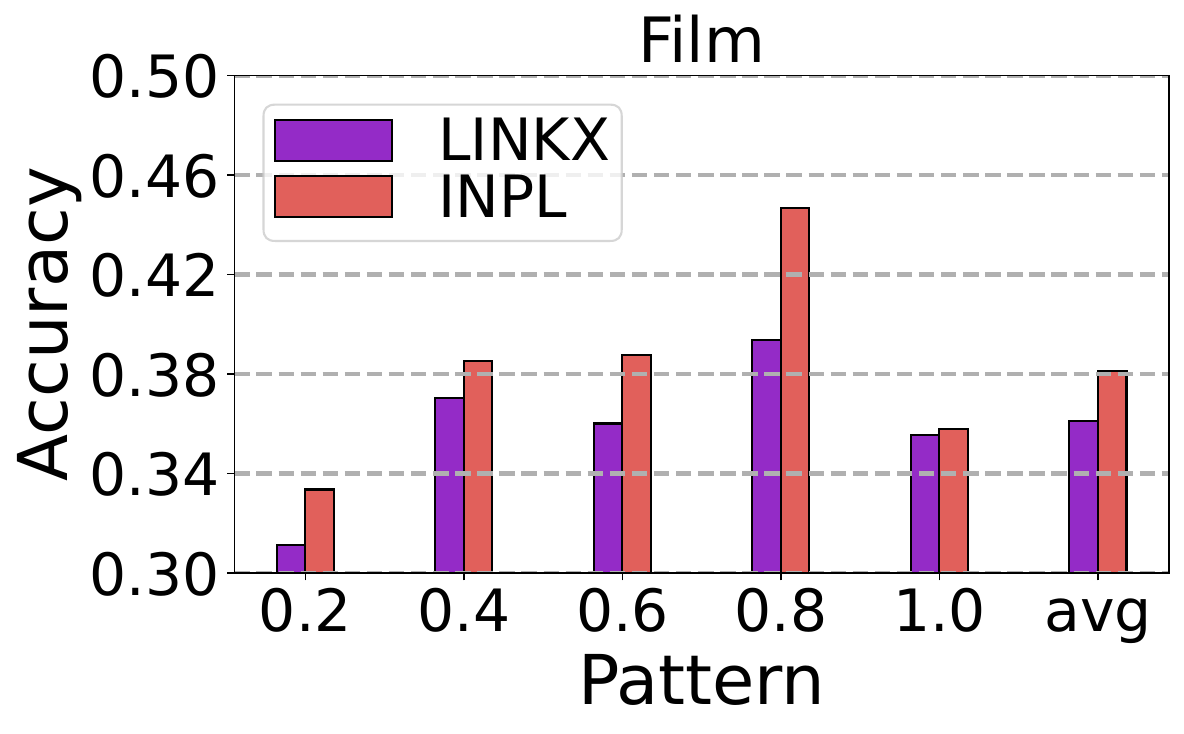}  
\end{subfigure}

\caption{Results of INPL and LINKX under {neighborhood pattern} distribution shifts for the task of semi-supervised node classification. Compared with LINKX, our method INPL improves the accuracy of node classification across different {neighborhood pattern distribution shifts} environments.}    
\label{figNeighbor1}       
\end{figure*}

\begin{figure*}
\centering
\begin{subfigure}{0.31\linewidth}
     \includegraphics[scale=0.22]{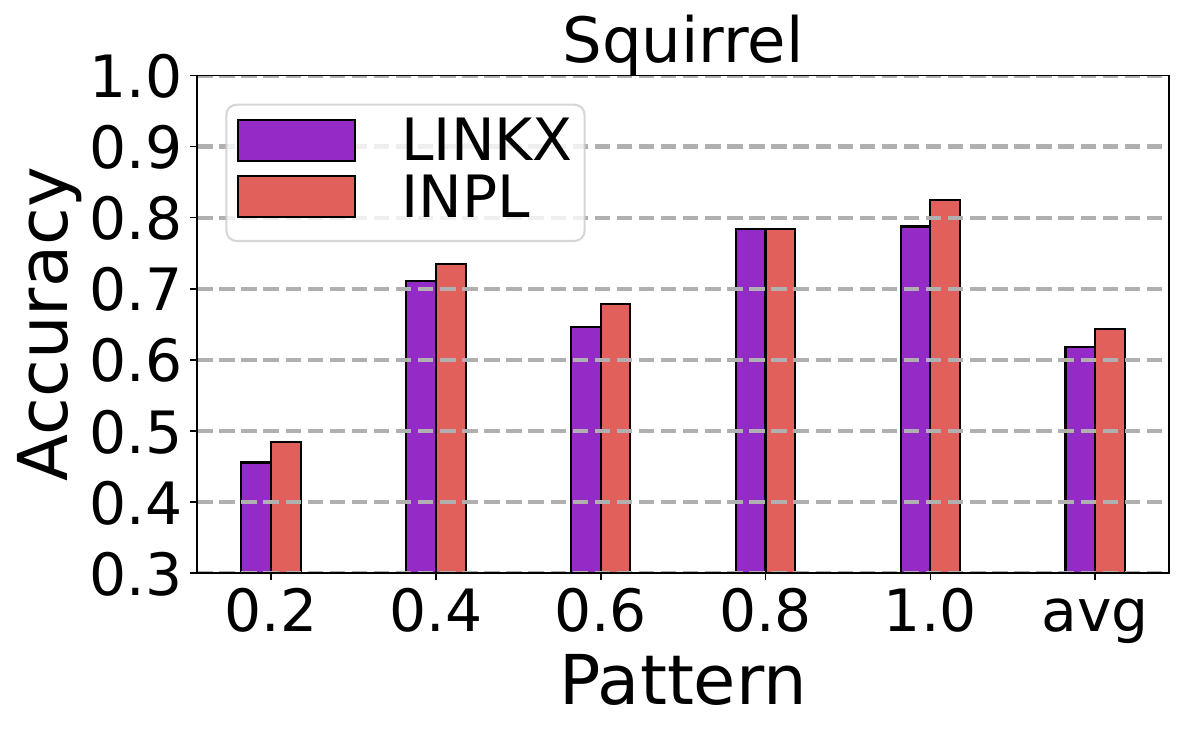}  
\end{subfigure}
\begin{subfigure}{0.31\linewidth}
     \includegraphics[scale=0.22]{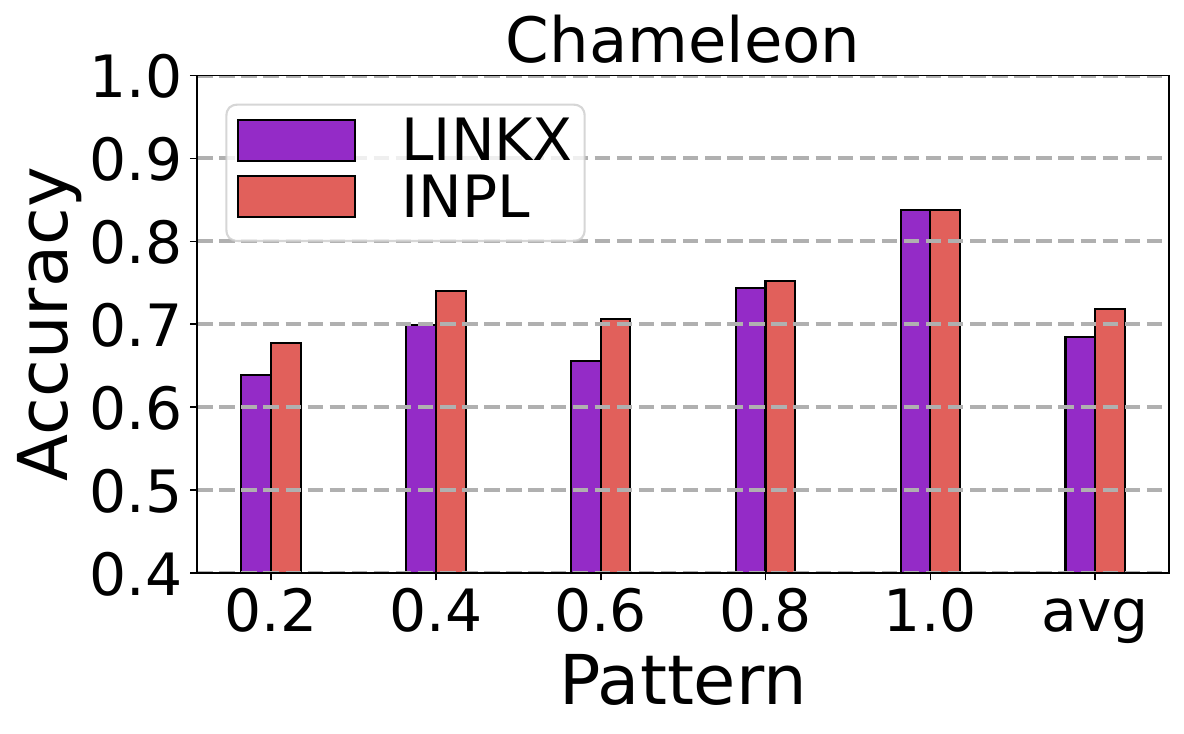}  
\end{subfigure}
\begin{subfigure}{0.31\linewidth}
     \includegraphics[scale=0.22]{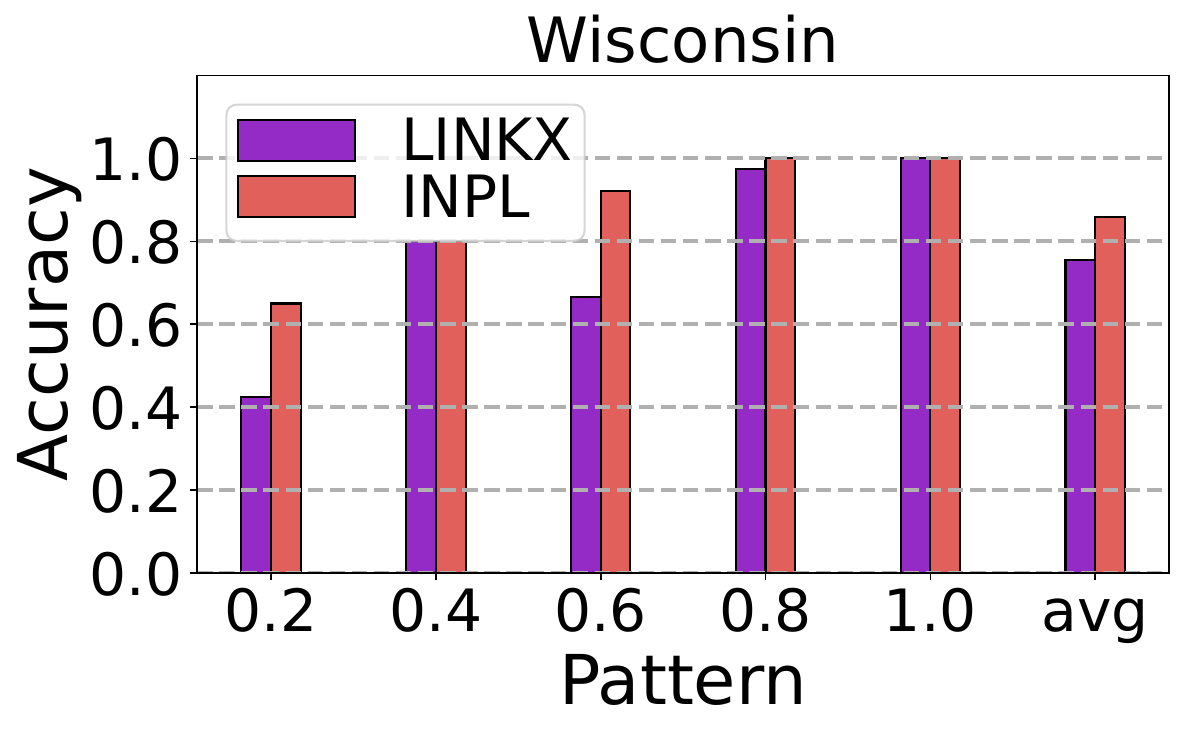}  
\end{subfigure}

\caption{Results of INPL and LINKX under {neighborhood pattern} distribution shifts for the task of semi-supervised node classification. Compared with LINKX, our method INPL improves the accuracy of node classification across different {neighborhood pattern }distribution shifts environments.}    
\label{figNeighbor2}       
\end{figure*}

\begin{figure*}
\centering
\begin{subfigure}{0.31\linewidth}
     \includegraphics[scale=0.22]{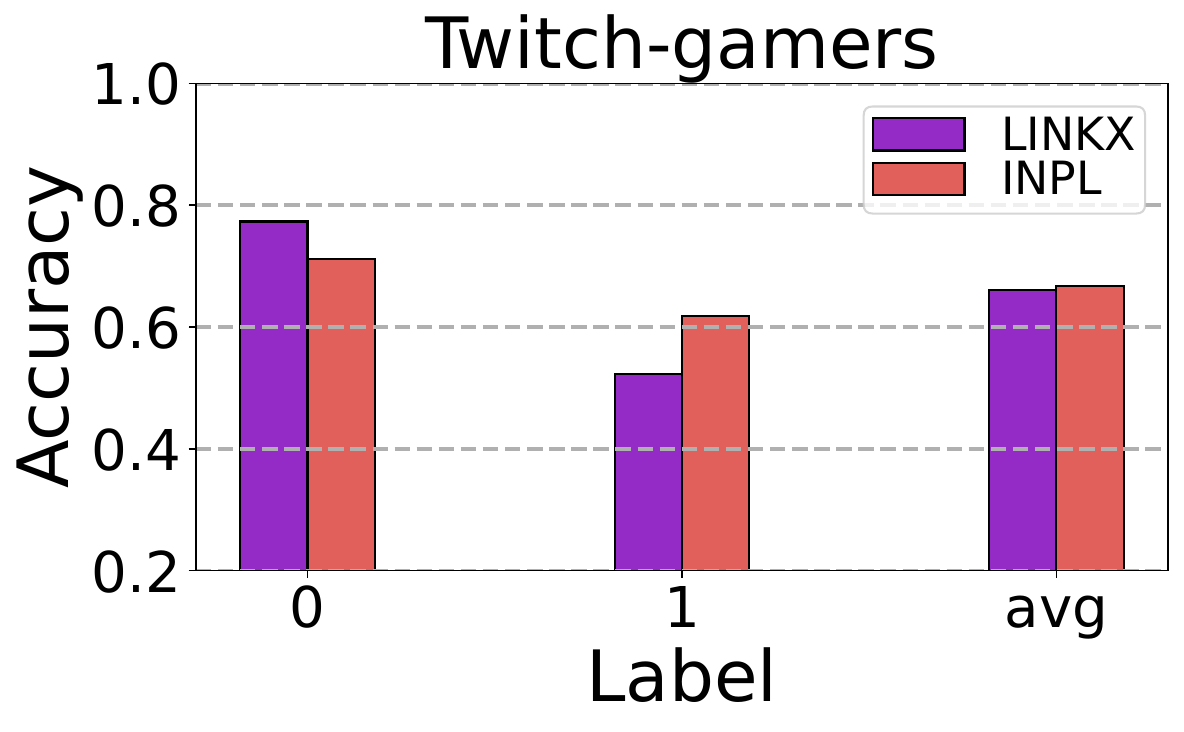}  
\end{subfigure}
\begin{subfigure}{0.31\linewidth}
     \includegraphics[scale=0.22]{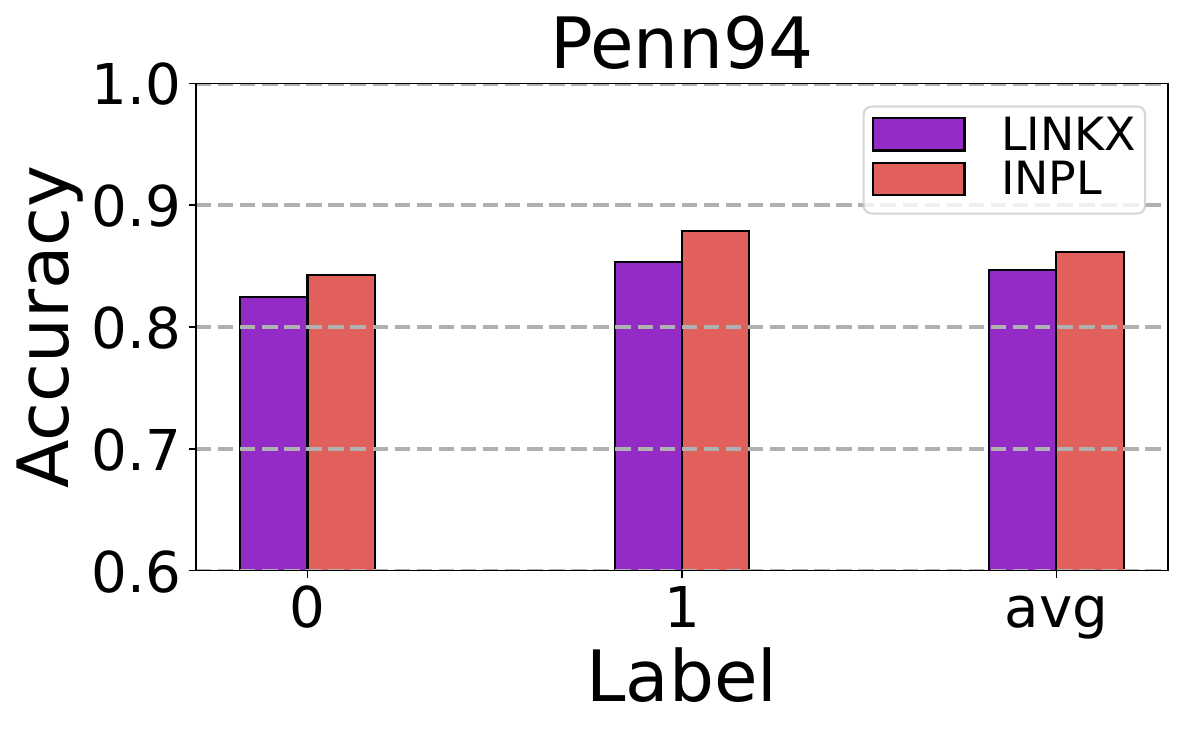}  
\end{subfigure}
\begin{subfigure}{0.31\linewidth}
     \includegraphics[scale=0.22]{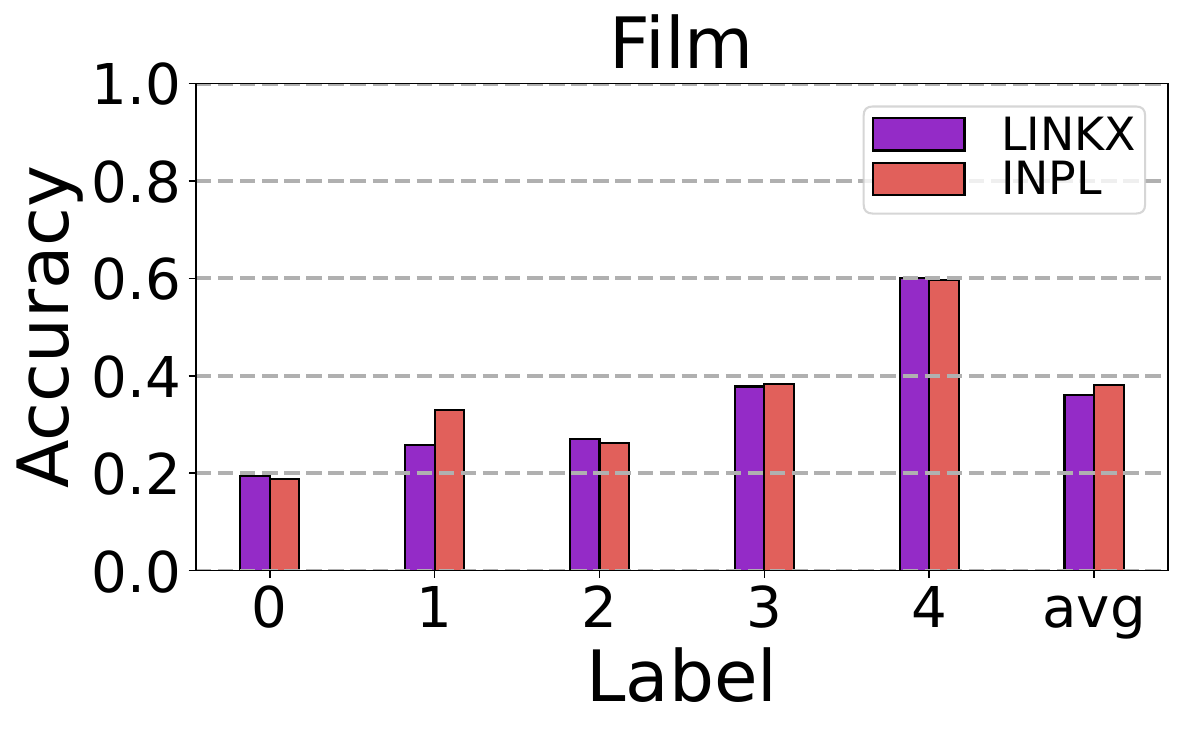}  
\end{subfigure}
\caption{Results of INPL and LINKX under label distribution shifts for the task of semi-supervised node classification on Twitch-gamers, Penn94 and Film. Compared with LINKX, our method INPL improves the accuracy of node classification across different label-biased environments.}    
\label{figLabel1}       
\end{figure*}
\section{Experiments}

In this section, we empirically evaluate the effectiveness of the proposed framework with a comparison to state-of-the-art graph neural networks on a wide variety of non-homophilous graph datasets. 


\subsection{Experiments Setup}
\textbf{Datasets.} We conduct experiments on eleven widely used non-homophilous datasets: film, squirrel, chameleon, cornell, texas wisconsin \cite{pei2020geom} , patents, pokec, genius, penn94 and twitch-gamers \cite{lim2021large}. We provide the detailed descriptions, statistics, and homophily measures of datasets in Appendix A.


\textbf{Baselines.} To evaluate the effectiveness of INPL, We compare INPL with the following representative semi-supervised learning methods:
(1) \textbf{Graph-agnostic methods}: Multi-Layer Perceptron (MLP) \cite{goodfellow2016deep}.
(2) \textbf{Node-feature-agnostic methods}: LINK \cite{zheleva2009join}.
(3) \textbf{Representative general GNNs}: GCN \cite{kipf2016semi}, GAT \cite{velivckovic2017graph}, jumping knowledge networks (GCNJK) \cite{xu2018representation} and APPNP \cite{gasteiger2018predict}.
(5) \textbf{Non-homophilous methods}: two H2GCN variants \cite{zhu2020beyond}, MixHop \cite{abu2019mixhop}, GPR-GNN \cite{chien2020adaptive}, GCNII \cite{chen2020simple}, three Geom-GCN variants \cite{pei2020geom}, LINKX \cite{lim2021large}. (4) \textbf{Debiased GNNs}: ImGAGN \cite{qu2021imgagn}, SL-DSGCN \cite{tang2020investigating},  BA-GNN \cite{chen2022ba}.

\textbf{Training and Evaluation.}We conduct experiments on pokec, snap-patents, genius, Penn94 and twitch-gamers with 50/25/25 train/val/test splits provided by \cite{lim2021large}, for datasets of film, squirrel, chameleon, cornell, texas and wisconsin, we follow the widely used semi-supervised setting in \cite{pei2020geom} with the standard 48/32/20 train/val/test splits. For each graph dataset, we report the mean accuracy with standard deviation on the test nodes in Table \ref{tab:results} with 5 runs of experiments. We use ROC-AUC as the metric for the class-imbalanced genius dataset (similar to \cite{lim2021large}), as it is less sensitive to class-imbalance than accuracy. For other datasets, we use classification accuracy as the metric.
 
\subsection{Overall Performance Comparison}
To evaluate the effectiveness of the proposed INPL framework, We conduct experiments on eleven non-homophilous graph datasets. Results are shown in Table \ref{tab:results}.
We have the following observations:
(1) Non-homophilous methods outperform graph-agnostic methods and representative general GNNs in all cases, the reason is that these representative GNNs rely on the homophily assumption, such assumption does not hold on non-homophilous graphs.
(2) INPL achieves state-of-the-art performance in most cases. 
INPL framework outperforms all non-homophilous methods in all these eleven non-homophilous datasets, the improvement is especially noticeable on small non-homophilous datasets of Cornell, Texas, and Wisconsin, with an improvement of \textbf{5.40\%}, \textbf{10.26\%} and \textbf{10.39\%}. 
The results demonstrate
that INPL could alleviate different biases on non-homophilous graphs, thus INPL outperforms in different biased environments and the overall classiﬁcation accuracy increases.
(3) INPL outperforms previous debiased GNNs. Previous debiased GNNs heavily rely on the homophilous assumption, which leads to poor generalization ability on non-homophilous graphs. The results demonstrate the effectiveness of INPL compared with previous debiased GNNs.

In summary, our INPL achieves superior performance on all these eleven non-homophilous datasets, which signiﬁcantly proves the effectiveness of our proposed framework and our motivation.

\subsection{Performance In Different Distribution Shifts}

We validate the effectiveness of our method under different distribution shifts.

\textbf{Neighborhood pattern distribution shifts}.
For neighborhood pattern distribution shifts test environments, we group the testing nodes of each graph according to their {neighborhood pattern}. {The neighborhood pattern distribution shifts between training and testing environments are various in this setting, where we evaluate methods in different environments rather than a single environment.} 
Results in Figure \ref{figNeighbor1} - \ref{figNeighbor2}
show that INPL achieves the best performance in all environments for all non-homophilous graphs, which evaluates INPL could alleviate neighborhood pattern-related bias.

\begin{figure*}
\centering
\begin{subfigure}{0.31\linewidth}
     \includegraphics[scale=0.22]{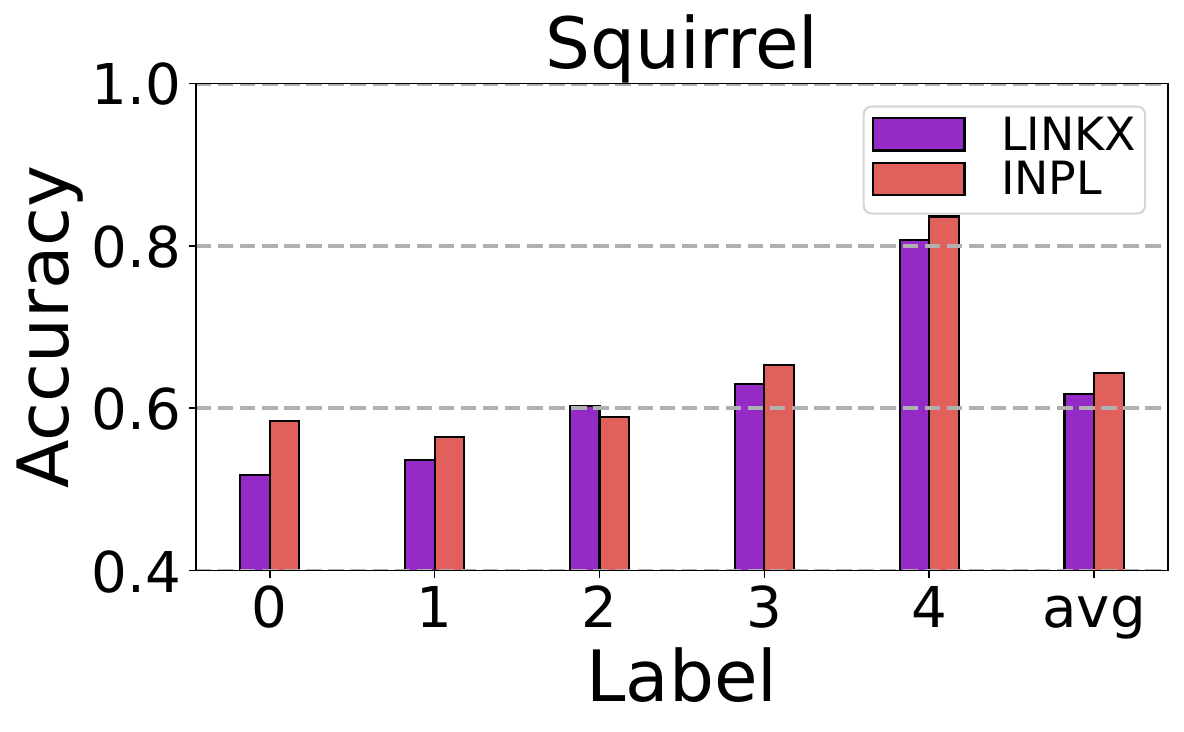}  
\end{subfigure}
\begin{subfigure}{0.31\linewidth}
     \includegraphics[scale=0.22]{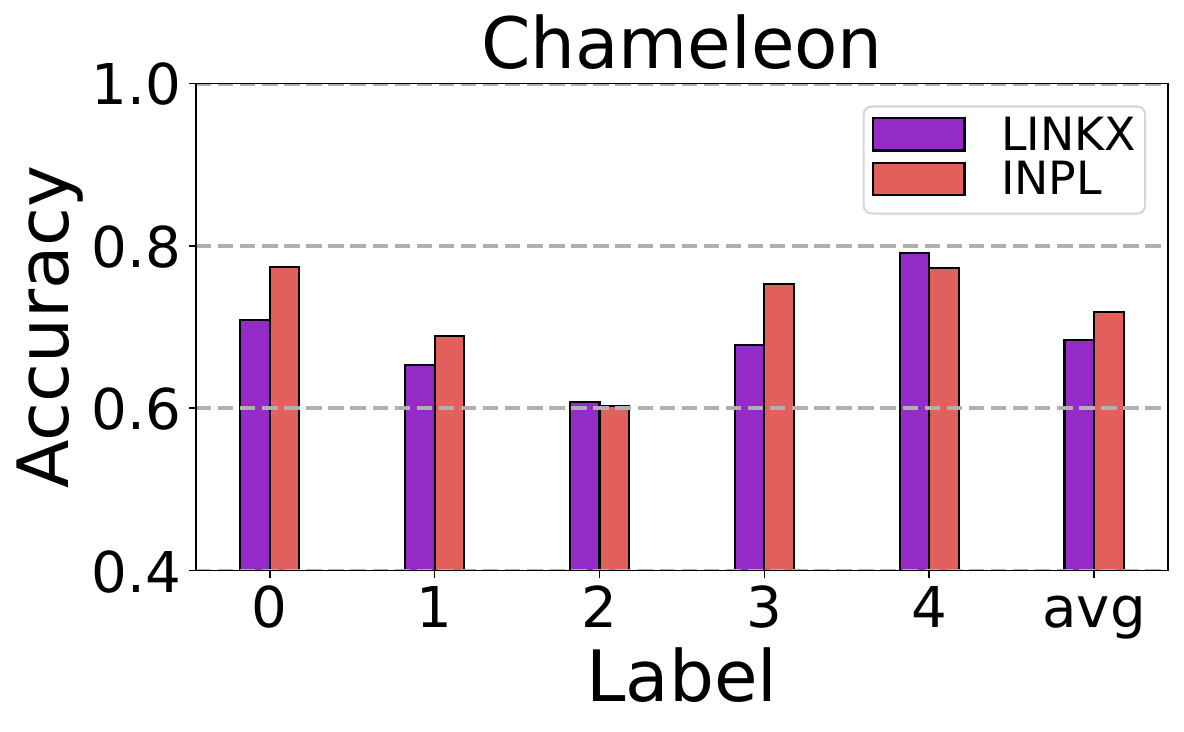}  
\end{subfigure}
\begin{subfigure}{0.31\linewidth}
     \includegraphics[scale=0.22]{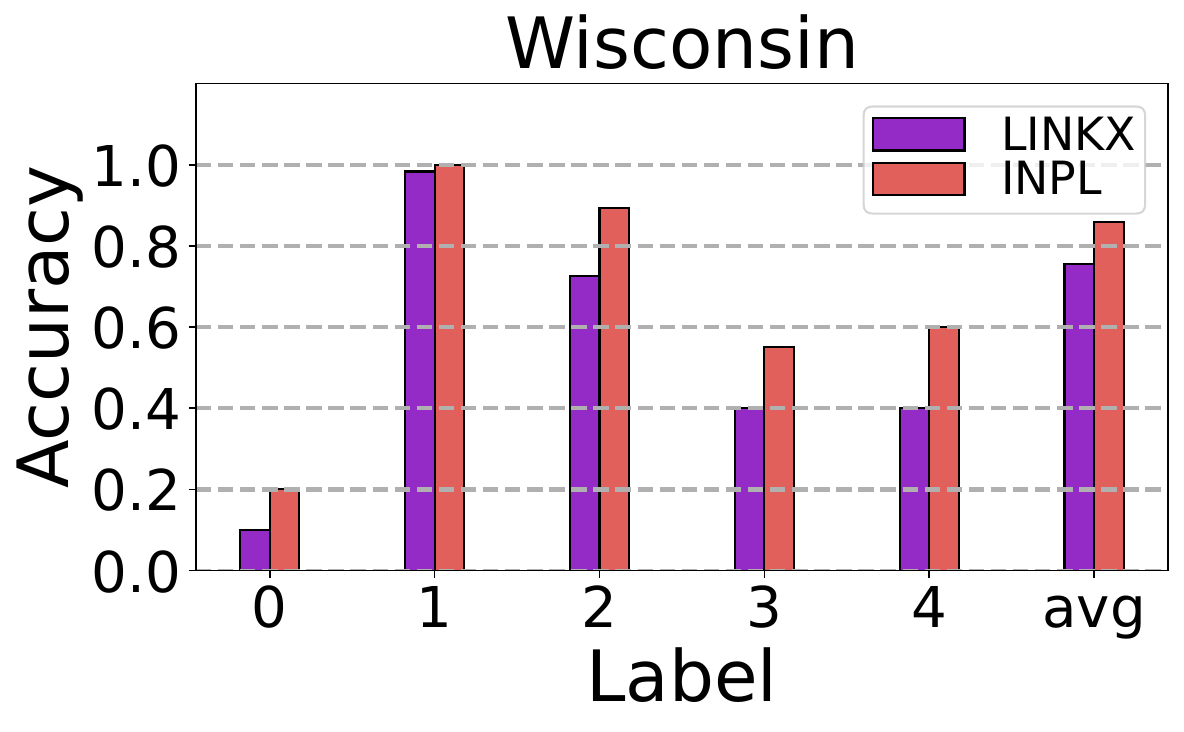}  
\end{subfigure}
\caption{Results of INPL and LINKX under label distribution shifts for the task of semi-supervised node classification. Compared with LINKX, 
INPL outperforms LINKX across different environments.
}    
\label{figLabel2}       
\end{figure*}

\textbf{Class distribution shifts}. For class distribution shifts, we group the testing nodes of each graph according to their labels. The distribution shifts between training and testing environments are various in this setting, where we evaluate methods in different environments rather than the average accuracy of all testing nodes in a single environment.
The results in class distribution shifts environments are shown in Figure \ref{figLabel1} - \ref{figLabel2}. Speciﬁcally, each ﬁgure in Figure \ref{figLabel1} - \ref{figLabel2} plots the evaluation results across different testing environments. From the results, we find that our INPL outperforms LINKX almost in all environments, which demonstrates the effectiveness of INPL. The reason why our INPL outperforms is that our method alleviates different biases based on the invariant representation, thus our method outperforms in different environments and the classification accuracy increases.

\begin{figure*}
\centering
\begin{subfigure}{0.31\linewidth}
     \includegraphics[scale=0.22]{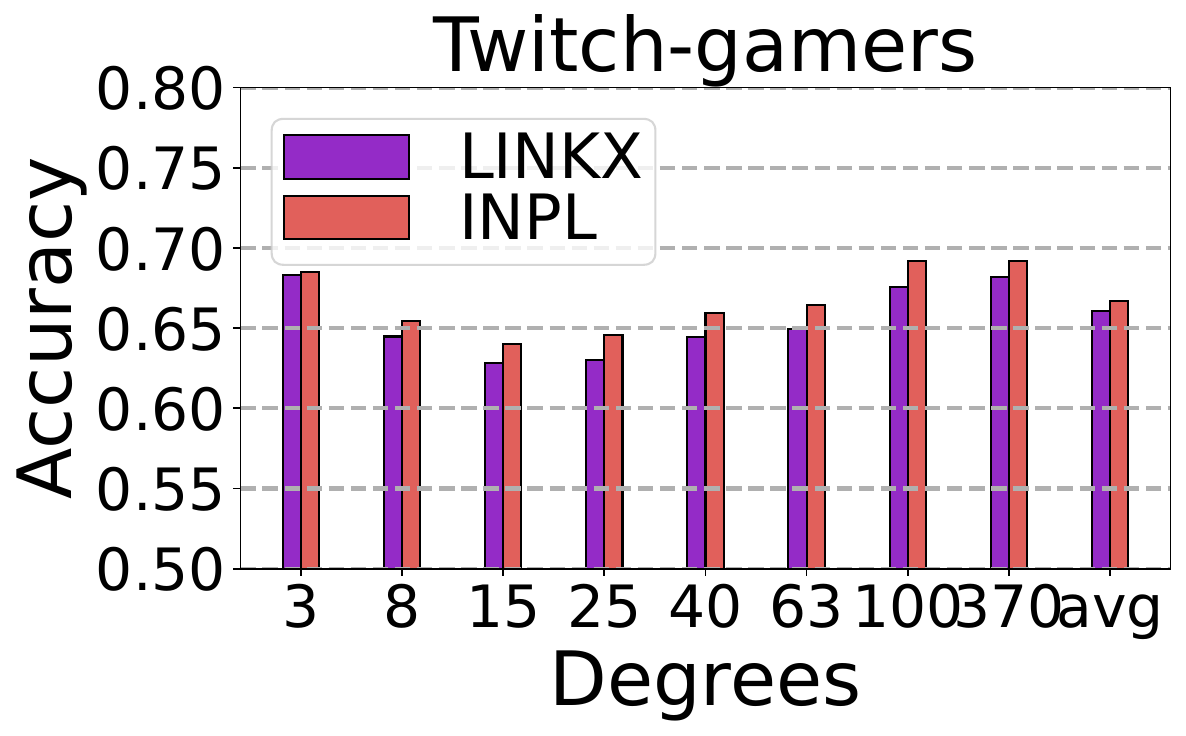}  
\end{subfigure}
\begin{subfigure}{0.31\linewidth}
     \includegraphics[scale=0.22]{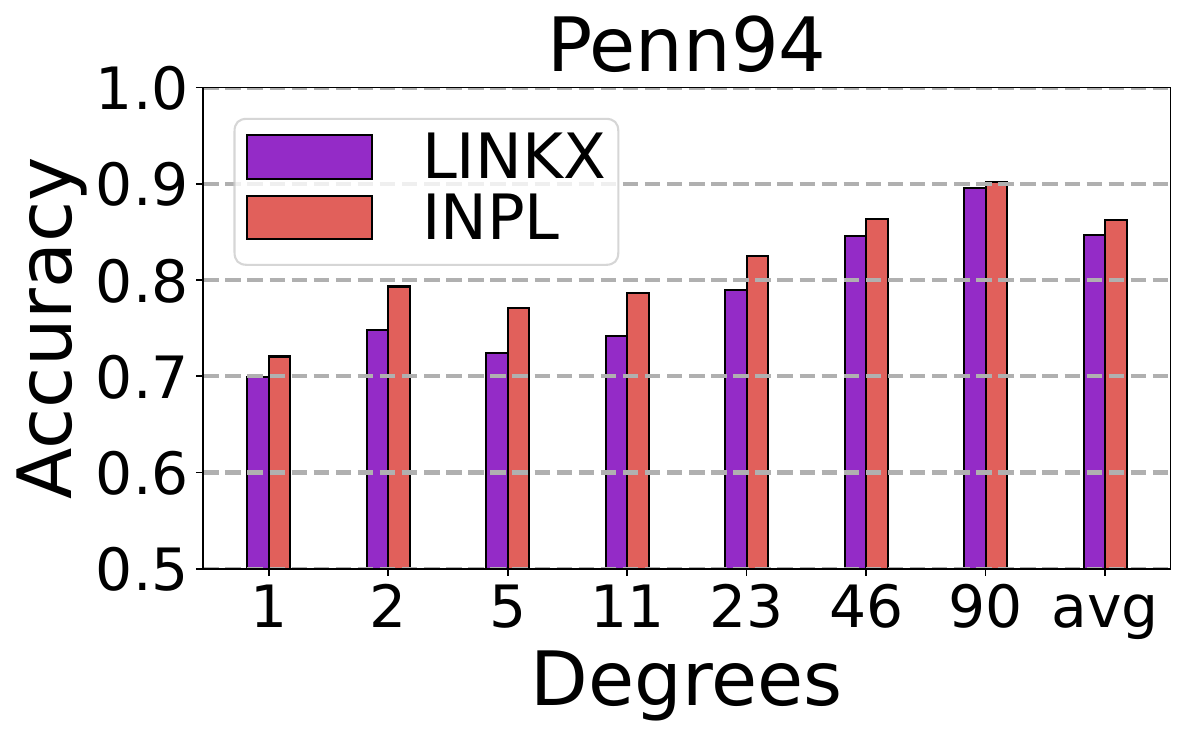}  
\end{subfigure}
\begin{subfigure}{0.31\linewidth}
     \includegraphics[scale=0.22]{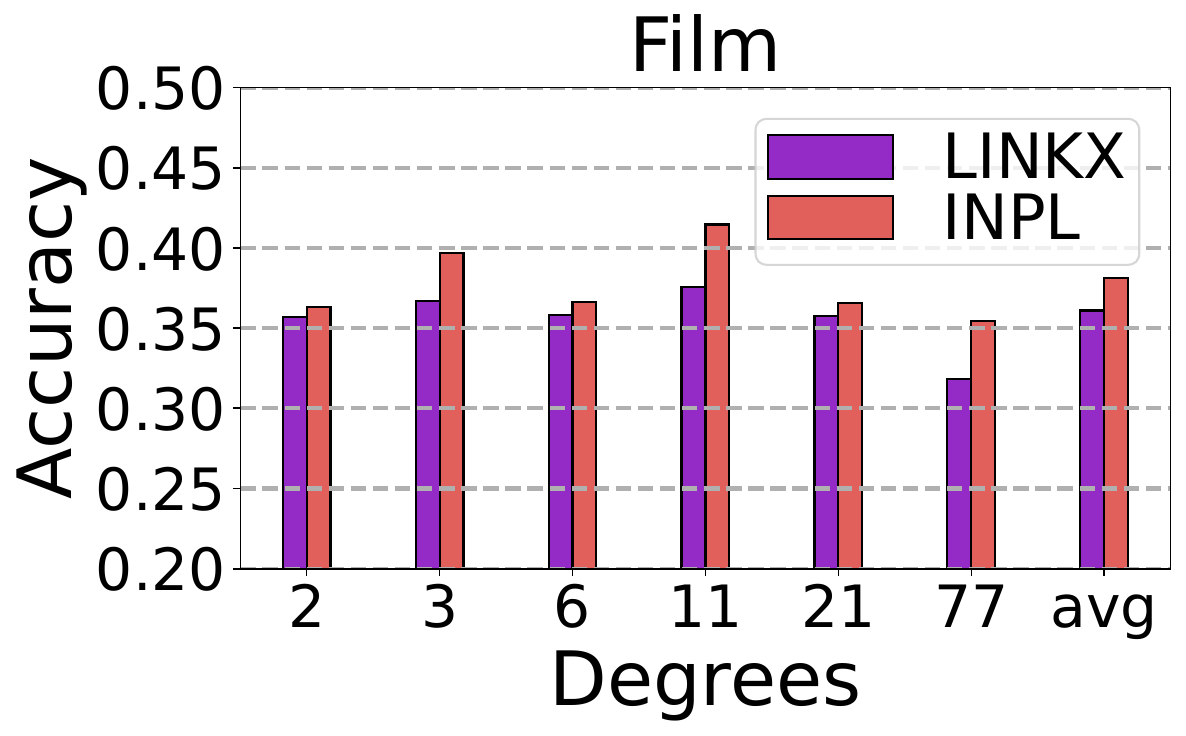}  
\end{subfigure}
\caption{Results of INPL and LINKX under degree distribution shifts for the task of semi-supervised node classification on Twitch-gamers, Penn94 and Film. Compared with LINKX, our method INPL improves the accuracy of node classification across different degree-biased environments.}    
\label{figdegree1}       
\end{figure*}

\textbf{Degree distribution shifts}. 
We organize each graph's testing nodes into different environments based on the degree.
In this setting, the distribution shifts between training and testing environments are various, and we
evaluate methods in different environments rather than the average accuracy of all testing nodes in a single environment.
Results in Figure \ref{figdegree1} show that INPL outperforms LINKX in all environments for all non-homophilous graphs, which evaluates INPL could alleviate degree-related bias. More experimental results of Squirrel, Chameleon and Wisconsin are in Appendix C.

\subsection{Performance in Unknown Environments}
\begin{figure*}
\centering
\begin{subfigure}{0.24\linewidth}
     \includegraphics[scale=0.17]{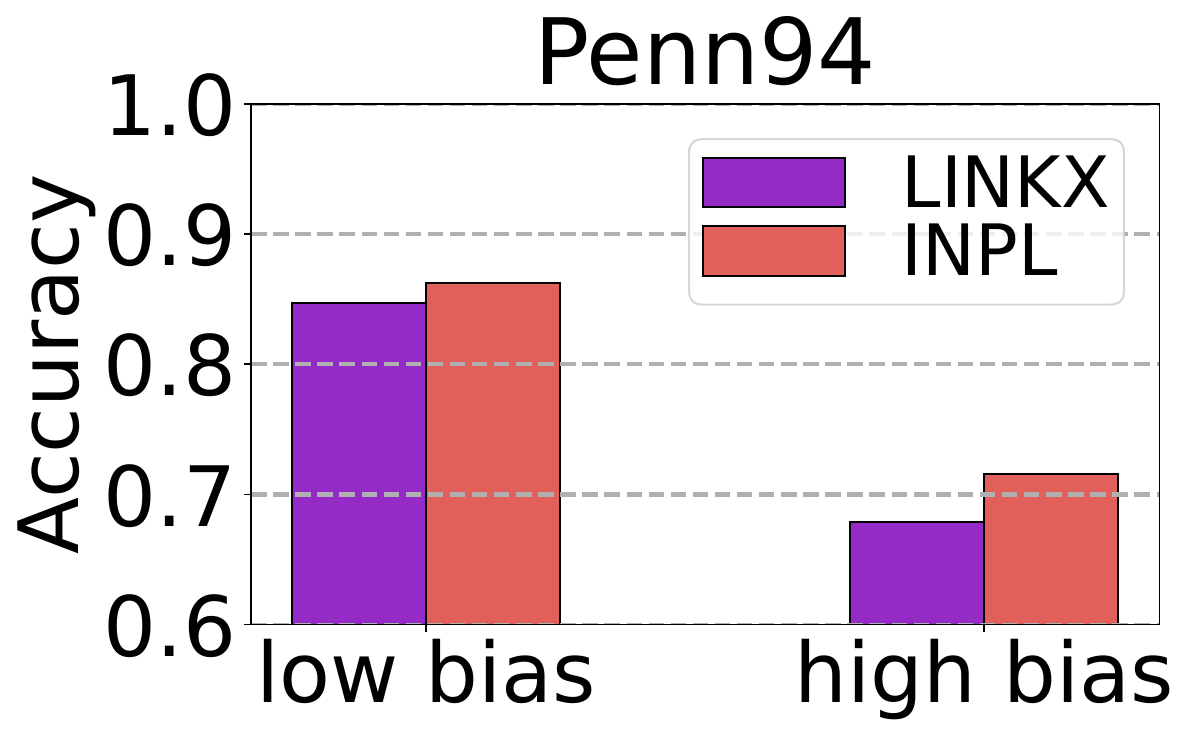}  
\end{subfigure}
\begin{subfigure}{0.24\linewidth}
     \includegraphics[scale=0.17]{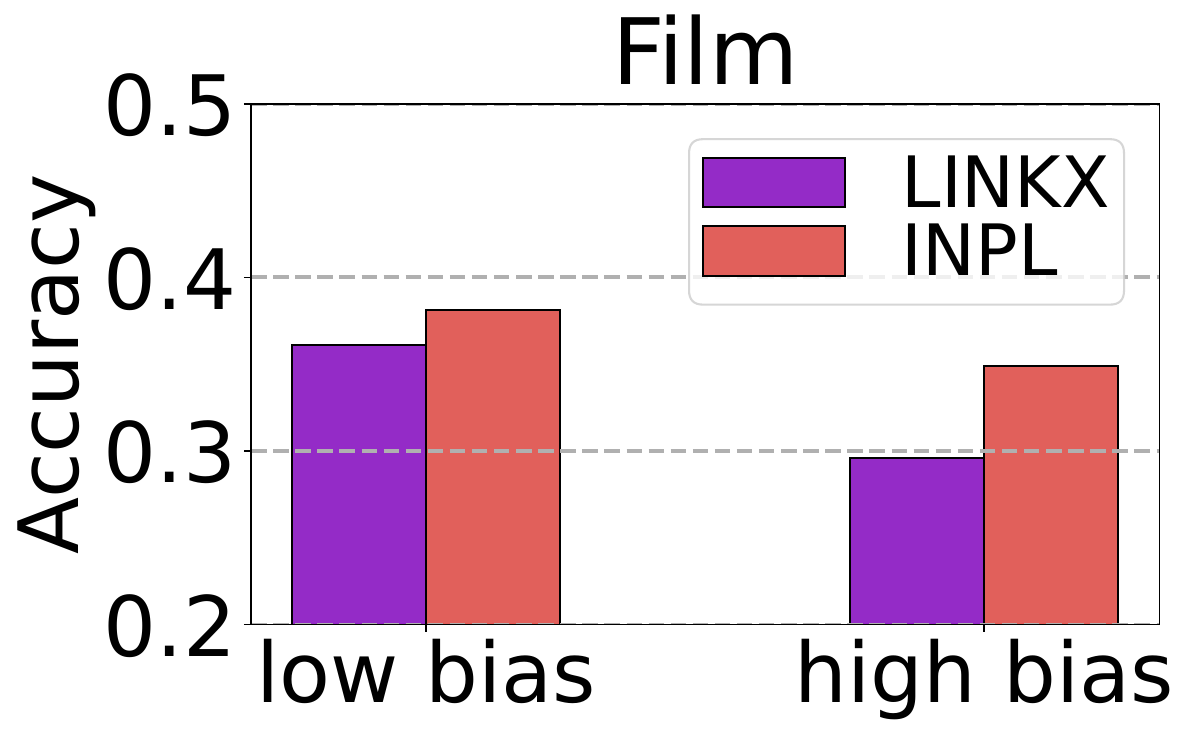}  
\end{subfigure}
\begin{subfigure}{0.24\linewidth}
     \includegraphics[scale=0.17]{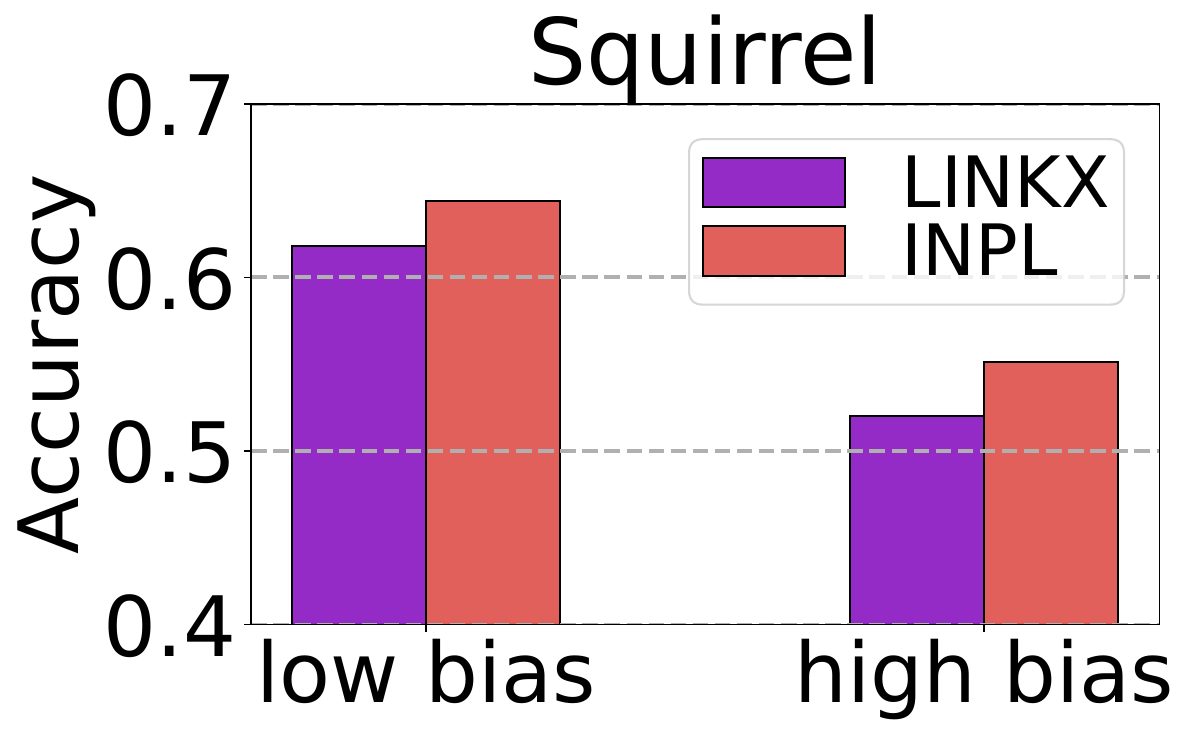}  
\end{subfigure}
\begin{subfigure}{0.24\linewidth}
     \includegraphics[scale=0.17]{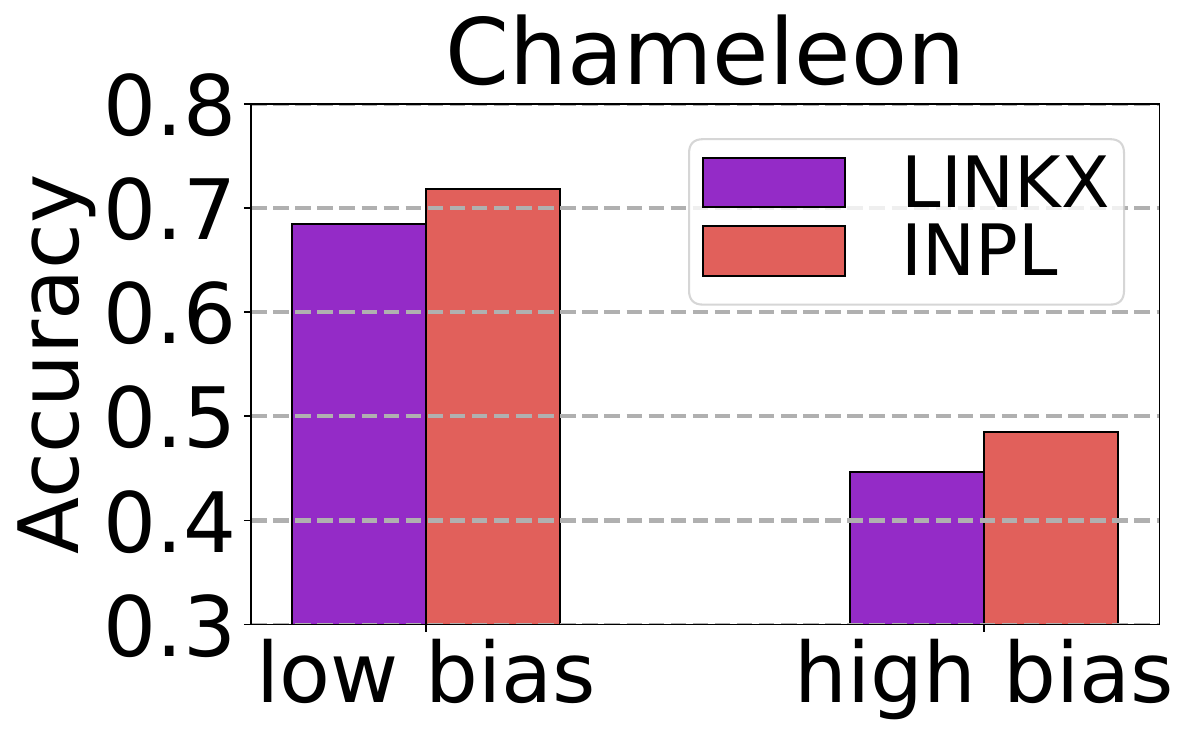}  
\end{subfigure}
\caption{Results of INPL and LINKX on strong distribution shifts for the task of semi-supervised node classification. INPL outperforms LINKX under strong distribution shifts, and the performance improvement in the high-bias case is much higher compared to the lower-bias case.}    
\label{fighighbias}       
\end{figure*}

\textbf{For degree-related high-bias environments}, We keep the nodes in the testing set unchanged and only select nodes in a specific small range of degrees as the training set, so there are strong distribution shifts between training and testing.
For low-bias environments, the training and test sets are the same as shown in Appendix C. The specific statistics and detailed experimental results of low-bias and high-bias environments are shown in Appendix C.
Results of low-bias and high-bias environments shown in Figure \ref{fighighbias} 
show that the performance of INPL and LINKX degrades under strong degree-related distribution shifts compared to the low-bias case, but the performance decline of INPL is much lower than LINKX,
which proves that INPL could alleviate degree-related distribution shifts in high-bias environments. Results of more baselines and additional datasets of twitch-gamers and pokec under degree-related high-bias environments are shown in Appendix C. More results of INPL and LINKX in different degree-related bias environments are shown in Appendix C. 
More experiments in {neighborhood pattern-related high-bias environments} are in Appendix C. 

\begin{figure*}
\centering
\begin{subfigure}{0.24\linewidth}
     \includegraphics[scale=0.23]{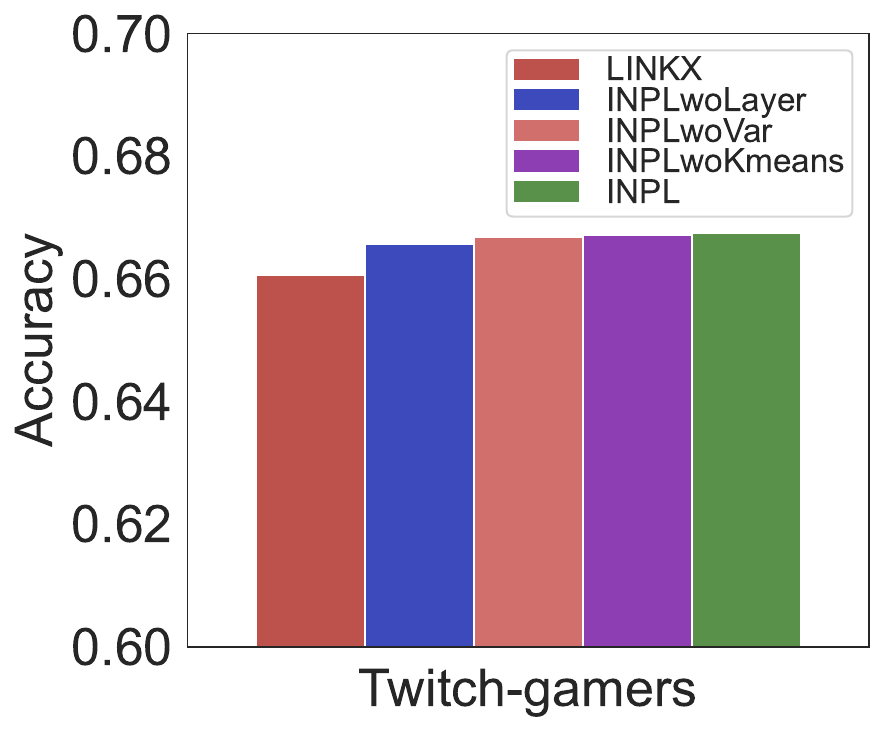}  
\end{subfigure}
\begin{subfigure}{0.24\linewidth}
     \includegraphics[scale=0.23]{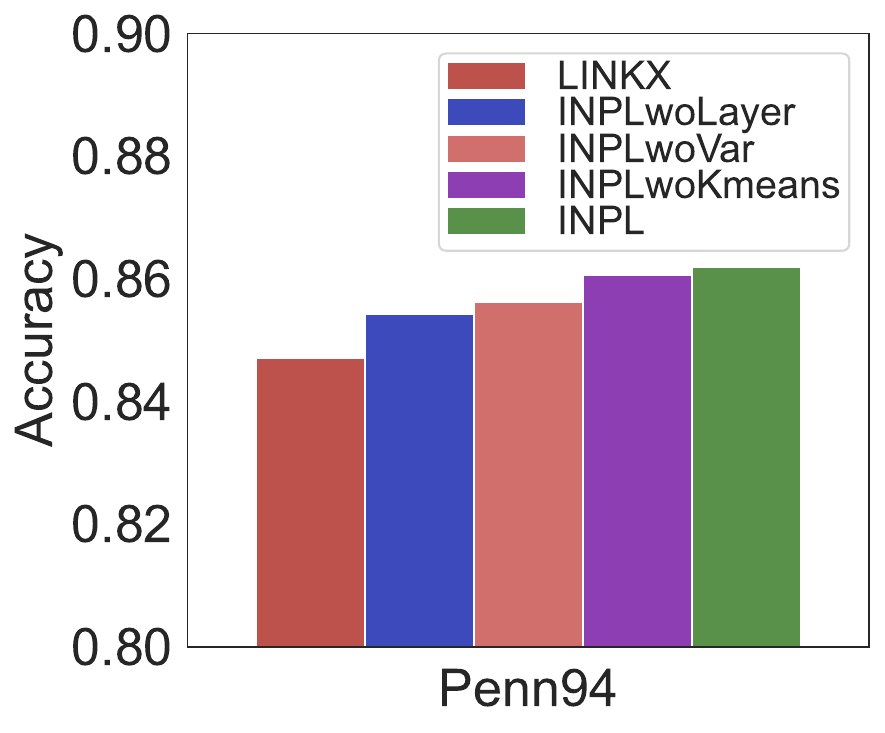}  
\end{subfigure}
\begin{subfigure}{0.24\linewidth}
     \includegraphics[scale=0.23]{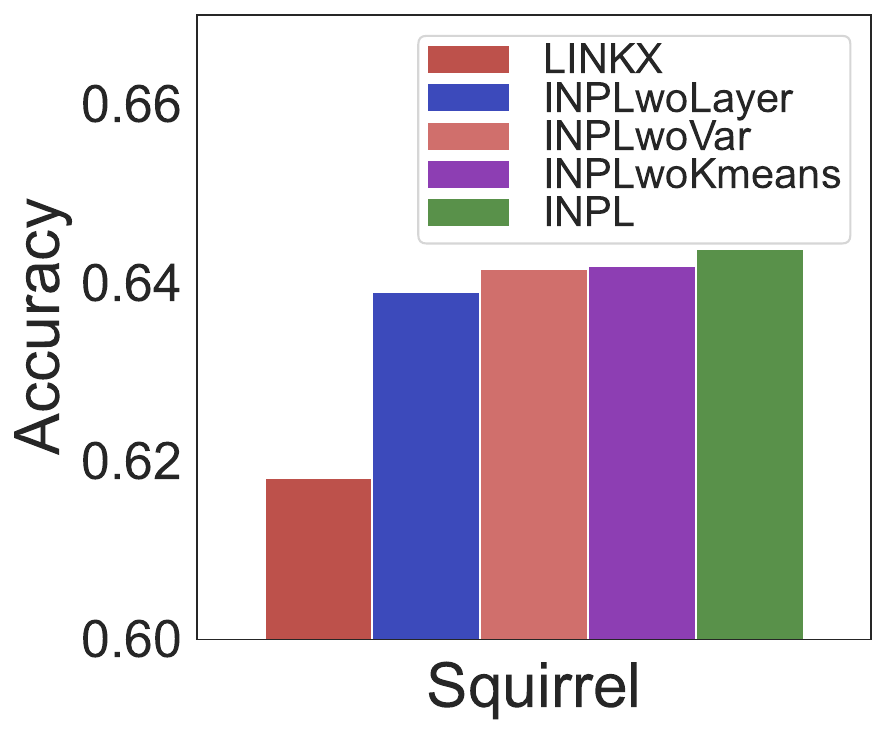}  
\end{subfigure}
\begin{subfigure}{0.24\linewidth}
     \includegraphics[scale=0.23]{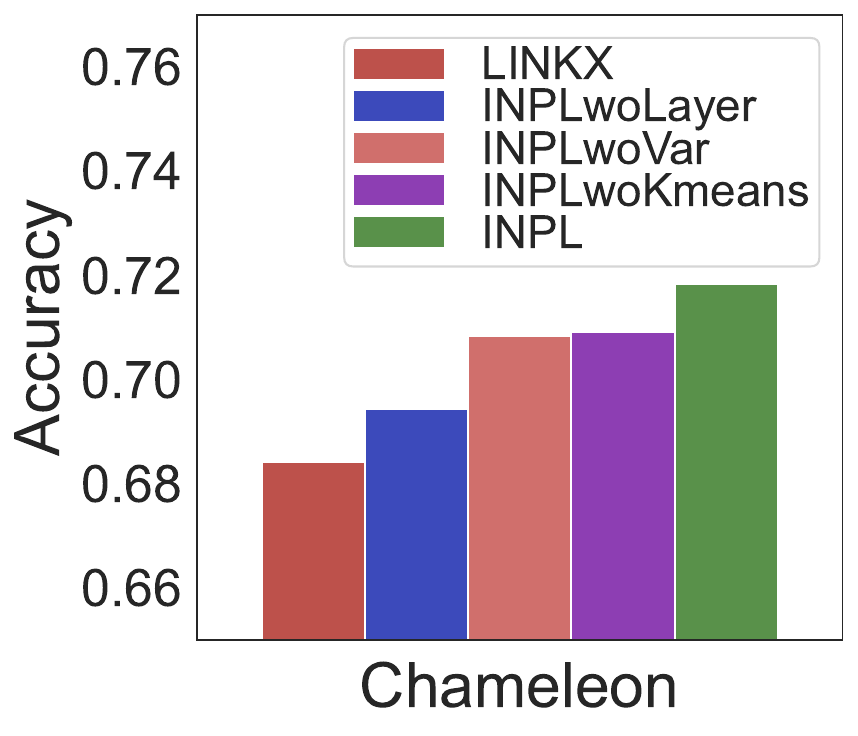}  
\end{subfigure}
\caption{Ablation study. Results indicate the removal of any module significantly decreases model performance. INPLwoLayer, INPLwoVar, and INPLwoKmeans still outperform LINKX.}    
\label{figablation}       
\end{figure*}



\subsection{Ablation Study}
In this section, we conduct an ablation study by changing parts of the entire framework. Speciﬁcally, we compare INPL with the following three variants:
(\textbf{INPLwoLayer}) {INPL without Adaptive Neighborhood Propagation Layer module.}
(\textbf{INPLwoVar}) {INPL without the variance penalty.}
(\textbf{INPLwoKmeans}) {using random graph partition instead of environment clustering.}

The results are shown in Figure \ref{figablation}. we find that there is a drop when we remove any proposed module of INPL. INPLwoLayer, INPLwoVar and INPLwoKmeans outperform LINKX, which demonstrates the effectiveness of other proposed modules. When we remove the Adaptive Neighborhood Propagation Layer module, INPLwoLayer has a significant drop, which illustrates the effectiveness of learning proper neighborhood information and invariant representation for each node. We also conduct an additional ablation study 
under degree-related high-bias environments in Appendix C.


\section{Conclusion}

In this paper, we focus on how to address the bias issue caused by the distribution shifts between training and testing distributions on non-homophilous graphs.
We propose a novel INPL framework that aims to learn invariant representation on non-homophilous graphs. 
To alleviate the neighborhood pattern distribution shifts problem on non-homophilous graphs, we propose Adaptive Neighborhood Propagation, where the Invariant Propagation Layer is proposed to combine both the high-order and low-order neighborhood information,  and Adaptive Propagation is proposed to capture the adaptive neighborhood information. 
To alleviate the distribution shifts in unknown test environments, we propose Invariant Non-Homophilous Graph Learning, which learns invariant graph representation on non-homophilous graphs. 
Extensive experiments on non-homophilous graphs validate the superiority of INPL and show that INPL could alleviate distribution shifts in different environments.






\bibliographystyle{plainnat}
\bibliography{neurips_2023}
\newpage

\input{appendix}

\end{document}

%% file: appendix.tex
\appendix

This is the Appendix for ``Discovering Invariant Neighborhood Patterns for Non-Homophilous Graphs''.

\begin{itemize}



\item Section ~\ref{sec:appendix_implementation_detail} reports dataset details.

\item Section ~\ref{sec:pseudo_code} shows the code and pseudo code of INPL.

\item Section ~\ref{sec:appendix_supplementary_exp} reports additional analysis of INPL.

\item Section ~\ref{discusstion} reports further discussion of this work.

\item Section ~\ref{broader_impact} reports broader impact of this work.

\end{itemize}

\section{Dataset Details}\label{sec:appendix_implementation_detail}

In this paper, we conduct experiments on eleven non-homophilous datasets, statistics of these eleven datasets are given in Table \ref{tab:datasets statistics}. To measure the homophily of a graph, the homophily ratio $h$ \cite{lim2021large} is given as:

\begin{linenomath}
\small
\begin{align}
h = \frac{1}{C-1}\sum_{k=0}^{C-1}\left[ h_k - \frac{\left|C_k\right|}{n}\right]_{+}
\label{Eq:h}
\end{align}
\end{linenomath}

where $\left[ a\right]_{+} = max(a,0)$, $C$ is the number of classes and $C_k$ denotes the set of nodes in class k, $k_u \in \{0, 1, ..., C - 1\}$ is the class label, $d_u$ is the number of neighbors of node $u$, and $d_u(k_u)$ is the number of neighbors of $u$ that have the same class label, $h_k$ is the class-wise homophily metric:
\begin{equation}
{
    h_k = \frac{\sum_{u\in C_k}d_u^{(k_u)}}{\sum_{u\in C_k}d_u}
}
\end{equation}

The homophily ratio $h$ measures the overall homophily level in the graph and thus we have $h \in [0, 1]$.
Speciﬁcally, graphs with higher $h$ tend to have more edges connecting nodes within the same class, or say stronger homophily, on the other hand, graphs with $h$ closer to 0 imply more edges connecting nodes in different
classes, that is, stronger non-homophily. In this paper, we focus on the graphs with low $h$ or say \emph{non-homophilous graphs}.

Here are the details of these eleven non-homophilous graph datasets in our experiments:
\begin{itemize}
    \item \emph{poekc} \cite{leskovec2005graphs} is the friendship graph of a Slovak online social network, where nodes are users and edges are directed friendship relations. Nodes are labeled with reported gender. 
    \item \emph{snap-patents} is a dataset of utility patents in the US. Each node is a patent, and edges connect patents that cite each other. Node features are derived from patent metadata.
    \item \emph{genius} \cite{chen2021deep} is a subset of the social network on genius.com — a site for crowdsourced annotations of song lyrics. Nodes are users, and edges connect users that follow each other on the site. The node features are user usage attributes like the Genius assigned expertise score, counts of contributions, and roles held by the user.
    \item \emph{twitch-gamers} \cite{rozemberczki2021twitch} is a connected undirected graph of relationships between accounts on the streaming platform Twitch. Each node represents a Twitch account, and edges exist between accounts that follow each other. The number of views, creation and update dates, language, life time, and whether the account is dead are all node features. The task of binary classification is to predict whether the channel contains explicit content.
    
    \item \emph{Penn94} \cite{traud2012social} is a friendship network from the 2005 Facebook 100 university student networks, where nodes represent students. Each node is labeled with the user's reported gender. Major, second major/minor, dorm/house, year, and high school are the features of the node.
    \item \emph{film/actor} is an actor co-occurrence network in which nodes represent actors and edges represent co-occurrence on the same Wikipedia page. It is the actor-only induced subgraph of the film-director-actor-writer network \cite{tang2009social}. The node feature vectors are a bag-of-words representation of the actors' Wikipedia pages' keywords. Each node is assigned one of five classes based on the topic of the actor's Wikipedia page.
    \item \emph{chameleon and squirrel} are Wikipedia networks \cite{rozemberczki2021multi}, in which nodes represent Wikipedia web pages and edges are mutual links between pages. And node features correspond to several informative nouns on the Wikipedia pages. Each node is assigned one of five classes based on the average monthly traffic of the web page.
    \item \emph{cornell, texas and wisconsin} are three subgraphs of the WebKB dataset collected by Carnegie Mellon University. Nodes represent web pages and edges are mutual links between pages. Node feature vectors are bag-of-word representation of the corresponding web pages. Each node is labeled with a student, project, course, staff, or faculty.
    
\end{itemize}

\begin{table*}[]
\captionsetup{font={small,stretch=1.25}, labelfont={bf}}
 \renewcommand{\arraystretch}{1}
    \caption{Statistics of the graph datasets. \#C is the number of distinct node classes, h is the homophily ratio.}
    \label{tab:datasets statistics}
    \centering
        \resizebox{0.99\linewidth}{!}{
    \begin{tabular}{cccccccccc}
        \toprule[1.5pt]
        \textbf{Dataset}      & \textbf{Nodes}& \textbf{Edges} & \textbf{Features} & \textbf{\#C}&  \textbf{Edge hom}& \textbf{h} & \textbf{\#Training Nodes} & \textbf{\#Validation Nodes}&  \textbf{\#Testing Nodes}  \\        \toprule[1.0pt]
        snap-patents & 2,923,922 & 13,975,788 & 269 & 5 & .073 & .100 & 50\% of nodes per class   & 25\% of nodes per class & Rest nodes\\
        
       pokec & 1,632,803 & 30,622,564 & 65 & 2 & .445 & .000 & 50\% of nodes per class   & 25\% of nodes per class & Rest nodes\\
       
    genius & 421,961 & 984,979 & 12 & 2 & .618 & .080 & 50\% of nodes per class   & 25\% of nodes per class & Rest nodes\\
    
	 Penn94 & 41554 & 1362229 & 5 & 2 & .470 & .046 & 50\% of nodes per class   & 25\% of nodes per class & Rest nodes\\

  twitch-gamers &  168114 & 6797557 & 7 & 2 & .545 & .090 & 50\% of nodes per class   & 25\% of nodes per class & Rest nodes\\
  
	 chameleon &  2277 & 36101  & 2325 & 5 & .23 & .062 & 48\% of nodes per class   & 32\% of nodes per class & Rest nodes\\
  
	 cornell &  183 & 295 & 1703 & 5 & .30 & .047 & 48\% of nodes per class   & 32\% of nodes per class & Rest nodes\\
  
	 film &  7600 & 29926 & 931 & 5  & .22 & .011 & 48\% of nodes per class   & 32\% of nodes per class & Rest nodes\\
  
      squirrel &  5201 & 216933  & 2089 & 5 & .22 & .025 & 48\% of nodes per class   & 32\% of nodes per class & Rest nodes\\
      
      texas &  183 & 309  & 1703 & 5 & .11 & .001 & 48\% of nodes per class   & 32\% of nodes per class & Rest nodes\\
      
      wisconsin &  251 & 499  & 1703 & 5 & .21 & .094 & 48\% of nodes per class   & 32\% of nodes per class & Rest nodes\\
      
        \toprule[1.5pt]    
    \end{tabular}}
\end{table*}

\section{Reproducible Code}
\label{sec:pseudo_code}

\textbf{Code}.
The code is public in: \\{https://github.com/AnnoymousCode/Invariant-Neighborhood-Pattern-Learning} 

\textbf{Pseudo Code of INPL}.
Algorithm~\ref{alg:pseudo_code} shows the pseudo-code.

\begin{algorithm}[t]
  \caption{Invariant Neighborhood Pattern Learning (INPL)}
  \label{alg:pseudo_code}
  \begin{algorithmic}[1]
    \State \textbf{Input:} Graph data $G=(A, X)$ and label $Y$.
    \While{Not converged or maximum epochs not reached} 
      \State Obtain multi-environment graph partition $G^{e}=(A^{e}, X^{e})$ via environment clustering module.
      \For{$e = 1$ to $|\mathcal{E}_{tr}|$}
        \State Compute loss $\mathcal{L}_p$ in Equation (4).  
        \State Compute loss $\mathcal{L}$ in Equation (2).
      \EndFor
      \State Optimize Invariant Propagation Network $\theta$ to minimize $\mathcal{L}_p$.
      \State Optimize Adaptive Propagation generator $\phi$ to minimize $\mathcal{L}$.
    \EndWhile
  \end{algorithmic}
\end{algorithm}

\section{Additional Analysis}\label{sec:appendix_supplementary_exp}

\subsection{Complexity analysis}
The time complexity of LINKX is $\mathcal{O}(d|E| + nd^2L)$, where $d$ is the hidden dimension, $L$ is the number of layers, $n$ is the number of nodes, and $|E|$ is the number of edges. 
INPL framework has several more propagation layers than LINKX (no more than 5 layers for all datasets), $d$, $|E|$ and $n$ are the same as in LINKX.
The number of layers is in the same order as in LINKX.
So the time complexity of INPL is the same as $\mathcal{O}(d|E| + nd^2L)$, where $L$ is the number of layers.

\subsection{LINKX: a strong, simple method}
Recently many non-homophilous methods have been proposed, such as MixHop \cite{abu2019mixhop}, H2GCN \cite{zhu2020beyond} and Geom-GCN \cite{pei2020geom}. However, these methods are not scalable, as they require more parameters and computational resources, which is not available for large graph datasets \cite{lim2021large}. Lim proposes a simple method LINKX \cite{lim2021large}, which overcomes the scalable problems and achieves state-of-the-art performance on large non-homophilous graph datasets. Specifically, LINKX separately embeds node features and adjacency information with MLPs, combines the embeddings by concatenation, then uses a ﬁnal MLP to generate predictions:
\begin{align}
    &h_A = MLP_A(A) \in R^{d\times n} \\
    &h_X = MLP_X(X) \in R^{d\times n} 
    \label{eq:hx}
\end{align}

\begin{align}
    Y = ML&P_f\left(\sigma\left(W[h_A;h_X]+h_A+h_X\right)\right)
\end{align}


 Despite the superior performance of LINKX compared with non-homophilous methods, 
 distribution shifts can significantly degrade the performance of LINKX since the testing data distribution can vary from training data distribution in real-world applications. We propose a novel Invariant Neighborhood Pattern Learning (INPL) to alleviate the distribution shifts problem on non-homophilous graphs. Our INPL is a scalable graph learning method, which outperforms LINKX and could achieve state-of-the-art performance for learning on large non-homophilous graphs by alleviating the distribution shifts problem on large non-homophilous graphs.

\subsection{Performance in degree-related bias environments}
\begin{figure*}[t]
\centering
\begin{subfigure}{0.31\linewidth}
     \includegraphics[scale=0.22]{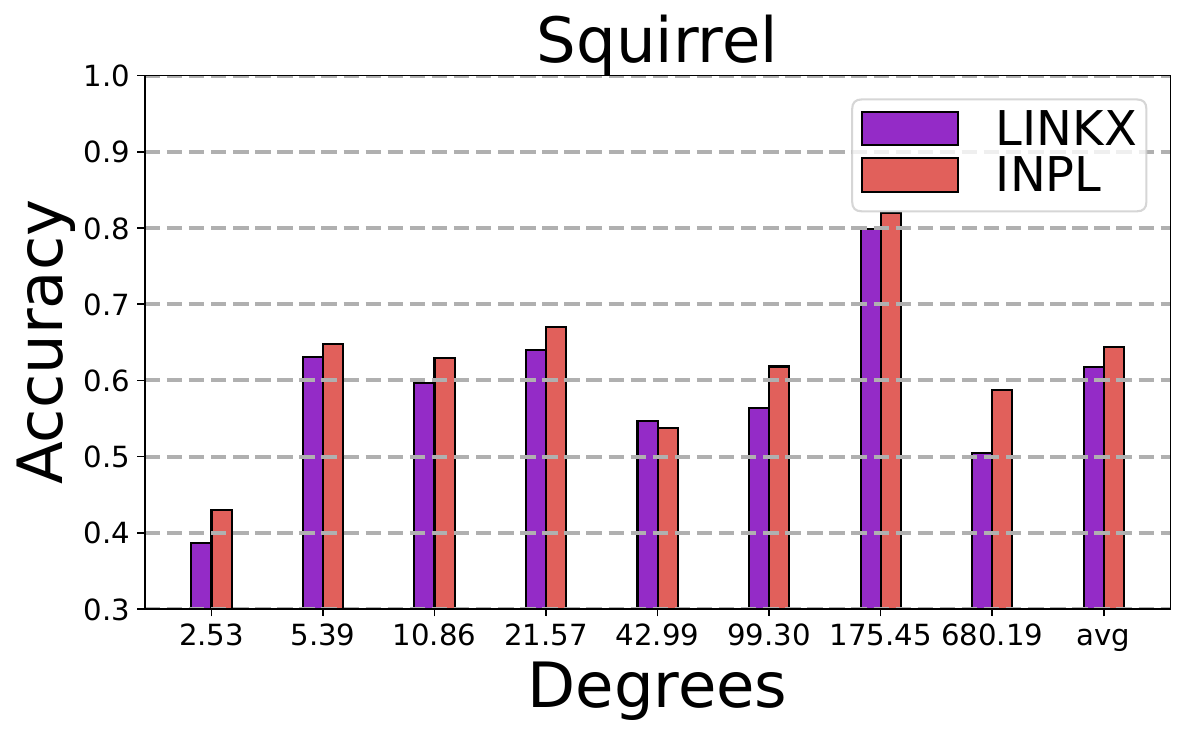}  
\end{subfigure}
\begin{subfigure}{0.31\linewidth}
     \includegraphics[scale=0.22]{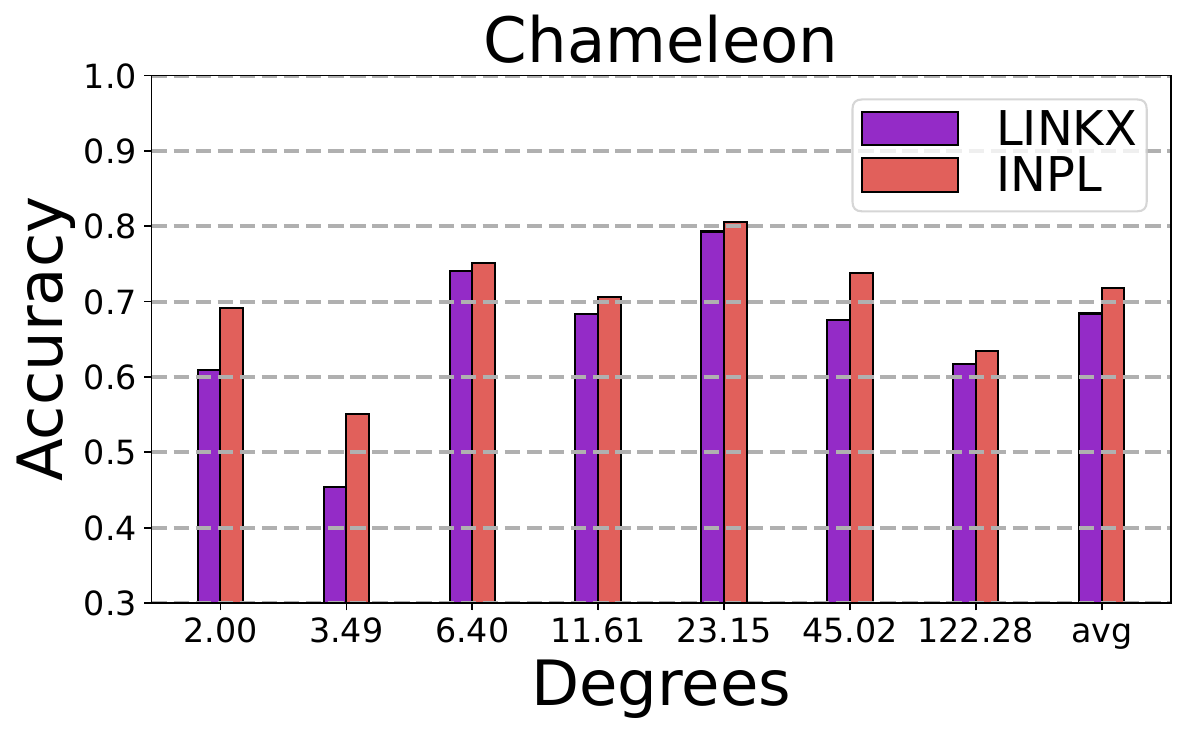}  
\end{subfigure}
\begin{subfigure}{0.31\linewidth}
     \includegraphics[scale=0.22]{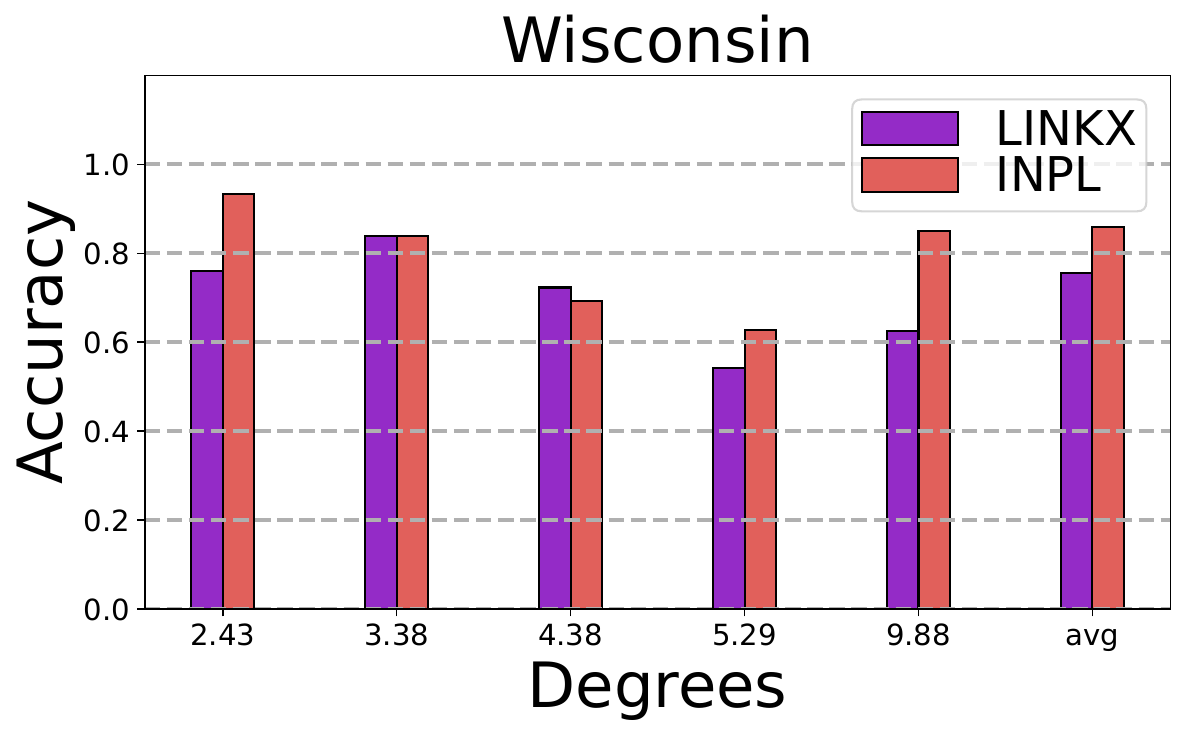}  
\end{subfigure}
\caption{Results of INPL and LINKX under degree distribution shifts for the task of semi-supervised node classification on Squirrel, Chameleon and Wisconsin. Compared with LINKX, our method INPL improves the accuracy of node classification across different degree-biased environments.}    
\label{figdegree2}       
\end{figure*}

We provide the detailed performance of INPL and LINKX in Squirrel, Chameleon and Wisconsin datasets.
Note that the main paper presents the results in Twitch-gamers, Penn94 and Film. 
And Figure \ref{figdegree2} shows the results in Squirrel, Chameleon and Wisconsin.

For degree-biased test environments, we group the testing nodes of each graph according to their degree. 
The distribution shifts between training and testing environments are various in this setting, where we evaluate methods in different environments rather than the average accuracy of all testing nodes in a single environment.

The results of Squirrel, Chameleon and  Wisconsin in degree-related bias
environments are shown in Figure \ref{figdegree2}. From the results, we find that our INPL outperforms LINKX almost in all biased environments, which demonstrates the effectiveness of INPL. The reason why our INPL outperforms is that our method alleviates different bias, thus our method outperforms in different biased environments and the classification accuracy increases.

\begin{table*}[]
\captionsetup{font={small,stretch=1.25}, labelfont={bf}}
 \renewcommand{\arraystretch}{1}
    \caption{Statistics and results of the degreee-related low-bias and high-bias environments of Penn94, Chameleon, Squirrel and Film.}
    \label{tab:bias statistics}
    \centering
        \resizebox{0.99\linewidth}{!}{
    \begin{tabular}{ccccccccc}
        \toprule[1.5pt]
	 Dataset& \multicolumn{2}{c}{Penn94} & \multicolumn{2}{c}{Chameleon} & \multicolumn{2}{c}{Squirrel} & \multicolumn{2}{c}{Film}  \\
  \toprule[1.0pt]
Biased environments & low-bias & high-bias & low-bias & high-bias & low-bias & high-bias & low-bias & high-bias \\
	 \#Train/Test graphs &  19407/9705 & 7509/9705 & 1092/456 & 373/456 & 2496/1041 & 2000/1041 & 3648/1520 & 2092/1520 \\
	 \#Nodes Train &  [1,4410] & [1,36]  & [2,733] & [2,8] & [2,1904] & [2,129] & [2,118] & [2,5] \\
	 \#Nodes Test &  [1,2613] & [1,2613] & [2,233] & [2,233] & [2,1320] & [2,1320] & [2,1304] & [2,1304] \\
   \toprule[1.0pt]
	 LINKX &  $84.71 \pm{0.52}$ & $67.90 \pm{0.79}$ & $68.42 \pm{1.38}$ & $44.65 \pm{1.26}$  & $61.81 \pm{1.80}$ & $52.03 \pm{1.15}$ & $36.10 \pm{1.55}$ & $29.59 \pm{2.02}$ \\
      \textbf{INPL} &  \bestcell$86.20 \pm{0.05}$ & \bestcell$71.60 \pm{1.90}$  & \bestcell$71.84 \pm{1.22}$ & \bestcell$48.51\pm{1.72}$ & \bestcell$64.38 \pm{0.62}$ & \bestcell$55.14 \pm{0.63}$ & \bestcell$38.12 \pm{0.36}$ &\bestcell $34.92 \pm{0.53}$ \\
        \toprule[1.5pt]    
    \end{tabular}}
\end{table*}

\begin{table*}[]
\captionsetup{font={small,stretch=1.25}, labelfont={bf}}
 \renewcommand{\arraystretch}{1}
    \caption{Statistics and results of the different biased environments of Penn94 dataset. We keep the nodes in the testing set unchanged and only select nodes in a specific small range of degrees as the training set, Same below.}
    \label{tab:datasets statistics1}
    \centering
        \resizebox{0.99\linewidth}{!}{
    \begin{tabular}{ccccccccc}
        \toprule[1.5pt]
 \#Train/Test graphs &  19407/9705 & 18009/9705 & 15521/9705 & 12067/9705 & 10480/9705 & 9007/9705 & 8506/9705 & 8015/9705 \\
	 \#Nodes Train &  [1,4410] & [1,160]  & [1,105] & [1,67] & [1,55] & [1,45] & [1,42] & [1,39] \\
	 \#Nodes Test &  [1,2613] & [1,2613] & [1,2613] & [1,2613] & [1,2613] & [1,2613] & [1,2613] & [1,2613]  \\
	 LINKX &  $84.71 \pm{0.52}$ & $83.84 \pm{0.33}$ & $81.99 \pm{0.31}$ & $77.81 \pm{0.33}$  & $74.98 \pm{0.55}$ & $71.52 \pm{0.52}$ & $70.36 \pm{0.67}$ & $69.41 \pm{0.84}$ \\
      
      \textbf{INPL} &\bestcell  $86.20 \pm{0.05}$ &\bestcell $85.31 \pm{0.30}$  &\bestcell $83.71 \pm{0.15}$ &\bestcell $79.95 \pm{0.48}$ &\bestcell $76.98 \pm{0.46}$ &\bestcell $74.39 \pm{1.53}$ &\bestcell $73.86 \pm{1.28}$ &\bestcell $72.91 \pm{1.66}$ \\
   \toprule[1.5pt]
	 \#Train/Test graphs &  7509/9705 & 6600/9705 & 5656/9705 & 4508/9705 & 4116/9705 & 3501/9705 & 2471/9705 & 1082/9705 \\
	 \#Nodes Train &  [1,36] & [1,31]  & [1,26] & [1,20] & [1,18] & [1,15] & [1,10] & [1,4] \\
	 \#Nodes Test &  [1,2613] & [1,2613] & [1,2613] & [1,2613] & [1,2613] & [1,2613] & [1,2613] & [1,2613]  \\
	 LINKX &  $67.90 \pm{0.79}$ & $65.90 \pm{1.01}$ & $64.04 \pm{1.18}$ & $62.86 \pm{0.55}$  & $61.95 \pm{0.73}$ & $60.74 \pm{0.53}$ & $58.98 \pm{0.50}$ & $58.45 \pm{0.24}$ \\
      
      \textbf{INPL} &\bestcell  $71.60 \pm{1.90}$ &\bestcell $69.52 \pm{2.42}$  &\bestcell $67.06 \pm{2.16}$ &\bestcell $65.60 \pm{3.03}$ &\bestcell $64.31 \pm{2.87}$ &\bestcell $62.70 \pm{2.03}$ &\bestcell $61.34 \pm{1.16}$ &\bestcell $61.50 \pm{1.19}$ \\
        \toprule[1.5pt]    
    \end{tabular}}
\end{table*}

\begin{table*}[]
\captionsetup{font={small,stretch=1.25}, labelfont={bf}}
 \renewcommand{\arraystretch}{1}
    \caption{Statistics and results of the different biased environments of Chameleon dataset.}
    \label{tab:datasets statistics2}
    \centering
        \resizebox{0.99\linewidth}{!}{
    \begin{tabular}{ccccccccc}
        \toprule[1.5pt]
\#Train/Test graphs &  1092/456 & 1003/456 & 901/456 & 842/456 & 781/456 & 718/456 & 690/456 & 622/456 \\
	 \#Nodes Train &  [2,733] & [2,80]  & [2,43] & [2,33] & [2,29] & [2,23] & [2,19] & [2,17] \\
	 \#Nodes Test &  [2,233] & [2,233] & [2,233] & [2,233] & [2,233] & [2,233] & [2,233] & [2,233]  \\
	 LINKX &  $68.42 \pm{1.38}$ & $68.46 \pm{0.59}$ & $64.43 \pm{0.57}$ & $62.94 \pm{1.89}$  & $62.81 \pm{1.54}$ & $56.58 \pm{0.52}$ & $55.35 \pm{0.59}$ & $51.58 \pm{1.15}$ \\
      
      \textbf{INPL} &\bestcell  $71.84 \pm{1.22}$ &\bestcell $71.36 \pm{0.53}$  &\bestcell $65.48 \pm{0.63}$ 
      &\bestcell $65.17 \pm{0.54}$ &\bestcell $65.02 \pm{0.68}$ &\bestcell $58.77 \pm{0.22}$ &\bestcell $56.71 \pm{0.51}$ &\bestcell $53.51 \pm{2.45}$ \\
   \toprule[1.5pt]
	 \#Train/Test graphs & 570/456 &  518/456 & 429/456 & 373/456 & 316/456 & 266/456 & 225/456  & 101/456 \\
	 \#Nodes Train &  [2,14] & [2,12] & [2,10]  & [2,8] & [2,7] & [2,6] & [2,5]  & [2,3] \\
	 \#Nodes Test &  [2,233] & [2,233] & [2,233] & [2,233] & [2,233] & [2,233] & [2,233] & [2,233]  \\
	 LINKX &  $50.61 \pm{1.24}$ & $47.81 \pm{0.77}$ & $48.12 \pm{2.99}$ & $44.65 \pm{1.26}$ & $40.40 \pm{1.14}$  & $36.45 \pm{1.22}$ & $32.32 \pm{0.96}$ &  $26.97 \pm{2.81}$ \\
      
      \textbf{INPL} &\bestcell  $51.62 \pm{1.24}$ &\bestcell $49.30 \pm{2.44}$  &\bestcell $49.87 \pm{0.33}$ &\bestcell $48.51 \pm{1.72}$ &\bestcell $42.24 \pm{0.33}$ &\bestcell $38.03 \pm{1.21}$ &\bestcell $35.27 \pm{0.57}$ &\bestcell $30.61 \pm{1.43}$ \\
        \toprule[1.5pt]    
    \end{tabular}}
\end{table*}

\begin{table*}[]
\captionsetup{font={small,stretch=1.25}, labelfont={bf}}
 \renewcommand{\arraystretch}{1}
    \caption{Statistics and results of the different biased environments of Squirrel dataset.}
    \label{tab:datasets statistics3}
    \centering
        \resizebox{0.99\linewidth}{!}{
    \begin{tabular}{ccccccccc}
        \toprule[1.5pt]
\#Train/Test graphs &  2496/1041 & 2450/1041 & 2400/1041 & 2361/1041 & 2306/1041 & 2235/1041 & 2000/1041 & 1900/1041 \\
	 \#Nodes Train &  [2,1904] & [2,813]  & [2,352] & [2,318] & [2,201] & [2,175] & [2,129] & [2,78] \\
	 \#Nodes Test &  [2,1320] & [2,1320] & [2,1320] & [2,1320] & [2,1320] & [2,1320] & [2,1320] & [2,1320]  \\
	 LINKX &  $61.81 \pm{1.80}$ & $61.13 \pm{1.25}$ & $60.92 \pm{0.63}$ & $61.48 \pm{0.59}$  & $60.33 \pm{0.49}$ & $58.52 \pm{0.54}$ & $52.03 \pm{1.15}$ & $49.68 \pm{0.69}$ \\
      
      \textbf{INPL} &\bestcell  $64.38 \pm{0.62}$ &\bestcell $63.53 \pm{0.34}$  &\bestcell $62.92 \pm{0.36}$ 
      &\bestcell $62.77 \pm{0.28}$ & \bestcell$62.36 \pm{0.59}$ &\bestcell $61.09 \pm{0.72}$ &\bestcell $55.14 \pm{0.63}$ &\bestcell $51.26 \pm{0.72}$ \\
   \toprule[1.5pt]
	 \#Train/Test graphs & 1697/1041 &  1489/1041 & 1194/1041 & 976/1041 & 764/1041 & 567/1041 & 459/1041  & 153/1041 \\
	 \#Nodes Train &  [2,39] & [2,26] & [2,16]  & [2,12] & [2,9] & [2,7] & [2,6]  & [2,3] \\
	 \#Nodes Test &  [2,1320] & [2,1320] & [2,1320] & [2,1320] & [2,1320] & [2,1320]  & [2,1320] & [2,1320]  \\
	 LINKX &  $46.75 \pm{0.52}$ & $43.88 \pm{0.32}$ & $40.63 \pm{0.27}$ & $37.25 \pm{0.45}$ & $32.55 \pm{0.26}$  & $31.09 \pm{0.22}$ & $28.84 \pm{0.74}$ &  $21.38 \pm{2.14}$ \\
      
      \textbf{INPL} &\bestcell  $48.59 \pm{0.99}$ &\bestcell $45.30 \pm{0.44}$  &\bestcell $41.92 \pm{0.13}$ &\bestcell $38.35 \pm{0.30}$ &\bestcell $33.76 \pm{0.57}$ &\bestcell $31.81 \pm{0.21}$ &\bestcell $30.11 \pm{0.47}$ &\bestcell $22.81 \pm{0.38}$ \\
        \toprule[1.5pt]    
    \end{tabular}}
\end{table*}

\begin{table*}[]
\captionsetup{font={small,stretch=1.25}, labelfont={bf}}
 \renewcommand{\arraystretch}{1}
    \caption{Statistics and results of the different biased environments of Film dataset.}
    \label{tab:datasets statistics4}
    \centering
        \resizebox{0.99\linewidth}{!}{
    \begin{tabular}{ccccccccc}
        \toprule[1.5pt]
\#Train/Test graphs &  3648/1520 & 3609/1520 & 3557/1520 & 3502/1520 & 3412/1520 & 3296/1520 & 3094/1520 & 2946/1520 \\
	 \#Nodes Train &  [2,118] & [2,49]  & [2,33] & [2,26] & [2,20] & [2,16] & [2,12] & [2,10] \\
	 \#Nodes Test &  [2,1304] & [2,1304] & [2,1304] & [2,1304] & [2,1304] & [2,1304] & [2,1304] & [2,1304]  \\
	 LINKX &  $36.10 \pm{1.55}$ & $35.97 \pm{2.18}$ & $35.36 \pm{1.66}$ & $35.52 \pm{1.80}$  & $35.37 \pm{1.29}$ & $35.51 \pm{1.43}$ & $34.99 \pm{0.95}$ & $35.06 \pm{1.79}$ \\
      
      \textbf{INPL} &\bestcell  $38.12 \pm{0.36}$ &\bestcell $37.60 \pm{0.83}$  &\bestcell $38.00 \pm{0.33}$ 
      &\bestcell $37.21 \pm{0.45}$ &\bestcell $37.39 \pm{0.29}$ &\bestcell $36.86 \pm{0.49}$ &\bestcell $36.55 \pm{0.17}$ &\bestcell $35.80 \pm{0.21}$ \\
   \toprule[1.5pt]
	 \#Train/Test graphs & 2832/1520 &  2680/1520 & 2525/1520 & 2321/1520 & 2092/1520 & 1789/1520 & 1363/1520  & 788/1520 \\
	 \#Nodes Train &  [2,9] & [2,8] & [2,7]  & [2,6] & [2,5] & [2,4] & [2,3]  & [2,2] \\
	 \#Nodes Test &  [2,1304] & [2,1304] & [2,1304] & [2,1304] & [2,1304] & [2,1304]  & [2,1304] & [2,1304]  \\
	 LINKX &  $32.80 \pm{1.83}$ & $32.43 \pm{4.95}$ & $33.29 \pm{1.99}$ & $28.46 \pm{7.17}$ & $29.59 \pm{2.02}$  & $28.84 \pm{1.05}$ & $28.92 \pm{3.57}$ &  $28.22 \pm{3.44}$ \\
      
      \textbf{INPL} &\bestcell  $35.79 \pm{0.40}$ &\bestcell $35.41 \pm{0.38}$  &\bestcell $35.60 \pm{0.47}$ &\bestcell $35.09 \pm{0.56}$ &\bestcell $34.92 \pm{0.53}$ &\bestcell $33.24 \pm{1.11}$ &\bestcell $32.12 \pm{1.50}$ &\bestcell $31.42 \pm{0.89}$ \\
        \toprule[1.5pt]    
    \end{tabular}}
\end{table*}

\begin{figure*}[t]
\centering
\begin{subfigure}{0.48\linewidth}
     \includegraphics[scale=0.20]{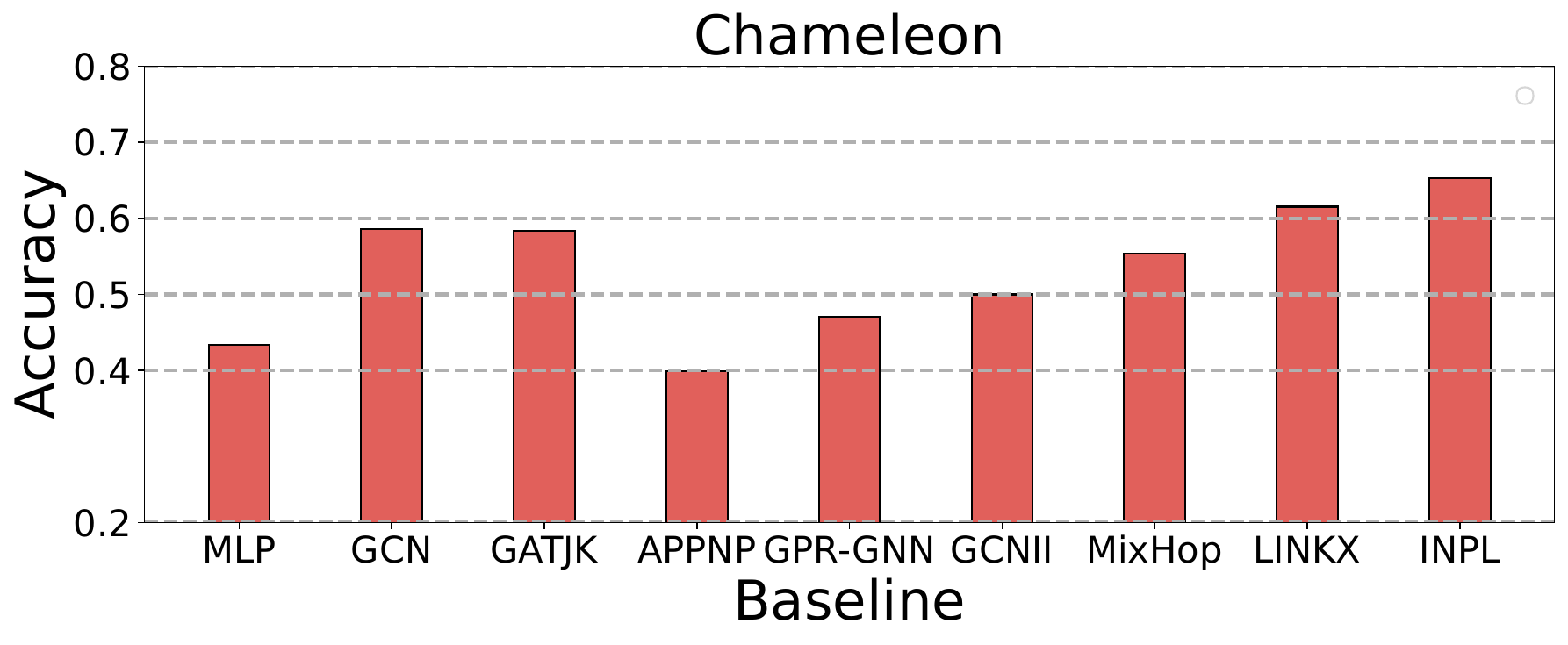}  
\end{subfigure}
\begin{subfigure}{0.48\linewidth}
     \includegraphics[scale=0.20]{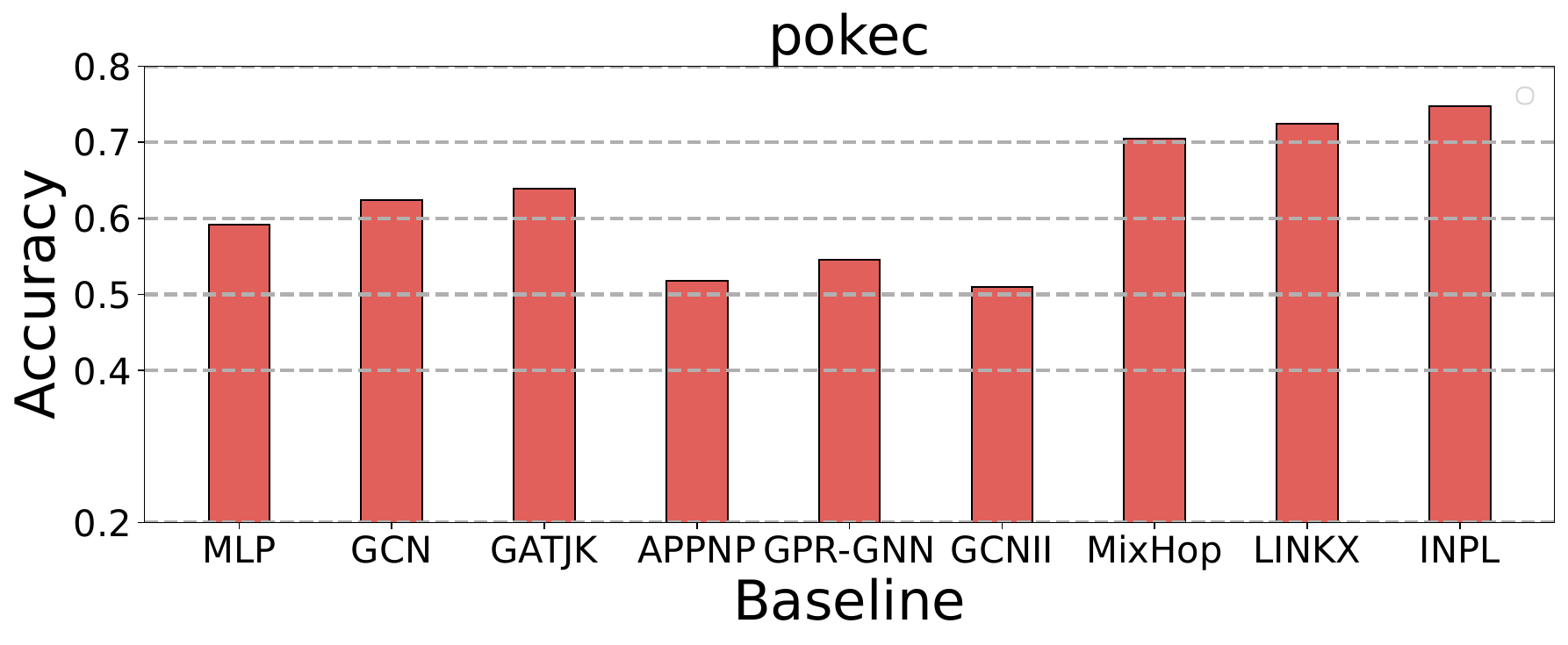}  
\end{subfigure}
\caption{Results of INPL and different baselines under neighborhood pattern-related high-bias environments for the task of semi-supervised node classification on chameleon and pokec. Compared with different baselines, our method INPL improves the accuracy of node classification under neighborhood pattern-related high-bias environments.}    
\label{figneighborhoodhighbias}       
\end{figure*}

\begin{table*}[]
\captionsetup{font={small,stretch=1.25}, labelfont={bf}}
 \renewcommand{\arraystretch}{1}
    \caption{Experimental results under degree-related high-bias environments. The best three results per dataset is highlighted. (M) denotes some (or all) hyperparameter settings run out of memory.}
    \label{tab:highbias-baselineresults}
    \centering
        \resizebox{0.99\linewidth}{!}{
    \begin{tabular}{ccccccccc}
        \toprule[1.5pt]
Dataset &  twitch-gamers & pokec & Penn94 & Film & Squirrel & Chameleon  \\
    \#Train/Test graphs &  60726/42029 & 324,428/408,160 & 7,509/9,705 & 2,092/1,520 & 2,000/1,041 & 373/456  \\
	 \#Nodes Train &  [1,67] & [1,8]  & [1,36] & [2,5] & [2,129] & [2,8]  \\
	 \#Nodes Test &  [1,35279] & [1,14854] & [1,2613] & [2,1304] & [2,1320] & [2,233]  \\
        
   \toprule[1.0pt]
	 MLP & $57.46 \pm{2.72}$ & $58.31 \pm{1.88}$ &\bestcell $69.72 \pm{0.04}$ & $31.46 \pm{0.06}$  & $29.15 \pm{0.13}$ & $37.06 \pm{0.22}$  \\
  GCN & $60.52 \pm{0.14}$ & $63.72 \pm{6.54}$ &\bestcell $69.69 \pm{1.78}$ & $28.62 \pm{0.77}$  & $48.18 \pm{0.52}$ & $44.30 \pm{3.30}$  \\
  GAT & (M) & (M) & (M) & $28.29 \pm{0.70}$  &\bestcell $51.09 \pm{1.36}$ & $45.04 \pm{1.32}$  \\
  GCNJK & $60.79 \pm{0.63}$ & $61.81 \pm{1.66}$ & $62.46 \pm{2.33}$ & $28.00 \pm{0.19}$  & $50.05 \pm{1.18}$ &\bestcell $45.75 \pm{1.85}$  \\
  GATJK & (M) & (M) & (M) & $27.11 \pm{0.74}$ & $47.84 \pm{0.54}$ &\bestcell $45.09 \pm{2.12}$  \\
  APPNP & $55.78 \pm{2.86}$ & $52.91 \pm{1.05}$ & $65.26 \pm{0.10}$ & $26.22 \pm{2.01}$  & $31.07 \pm{0.19}$ & $40.48 \pm{0.63}$  \\
  H\textsubscript{2}GCN-1 & (M) & (M) & (M) & $28.04 \pm{0.14}$  & $30.68 \pm{0.69}$ & $38.82 \pm{1.06}$  \\
  H\textsubscript{2}GCN-2 & (M) & (M) & (M) &\bestcell $32.16 \pm{0.88}$  & $31.10 \pm{0.46}$ & $37.06 \pm{1.89}$  \\
  MixHop &\bestcell $60.85 \pm{0.35}$ &\bestcell $69.56 \pm{1.24}$ & $68.91 \pm{0.75}$ &\bestcell $33.24 \pm{0.77}$  & $40.40 \pm{1.44}$ & $43.68 \pm{2.20}$  \\
  GPR-GNN & $57.11 \pm{3.74}$ & $52.33 \pm{1.82}$ & $68.00 \pm{0.15}$ & $32.20 \pm{0.56}$ & $33.72 \pm{0.24}$ & $43.38 \pm{0.29}$ \\
  
GCNII & $58.26 \pm{0.42}$ & $50.78 \pm{0.22}$ & $64.46 \pm{0.64}$ & $28.38 \pm{0.76}$  & $32.72 \pm{0.16}$ & $42.68 \pm{2.37}$  \\
LINKX &\bestcell $65.06 \pm{0.12}$ &\bestcell $70.42 \pm{0.97}$ & $67.90 \pm{0.79}$ & $29.59 \pm{2.02}$  &\bestcell $52.03 \pm{1.15}$ & $44.65 \pm{1.26}$  \\

\textbf{INPL(Ours)} &\bestcell  $66.14 \pm{0.04}$ &\bestcell  $74.26 \pm{0.18}$ &\bestcell  $71.60 \pm{1.90}$ &\bestcell  $34.92 \pm{0.53}$  &\bestcell  $55.14 \pm{0.63}$ &\bestcell  $48.51 \pm{1.72}$  \\
  
        \toprule[1.5pt]    
    \end{tabular}}
\end{table*}

\begin{figure*}[t]
\centering
\begin{subfigure}{0.96\linewidth}
     \includegraphics[scale=0.40]{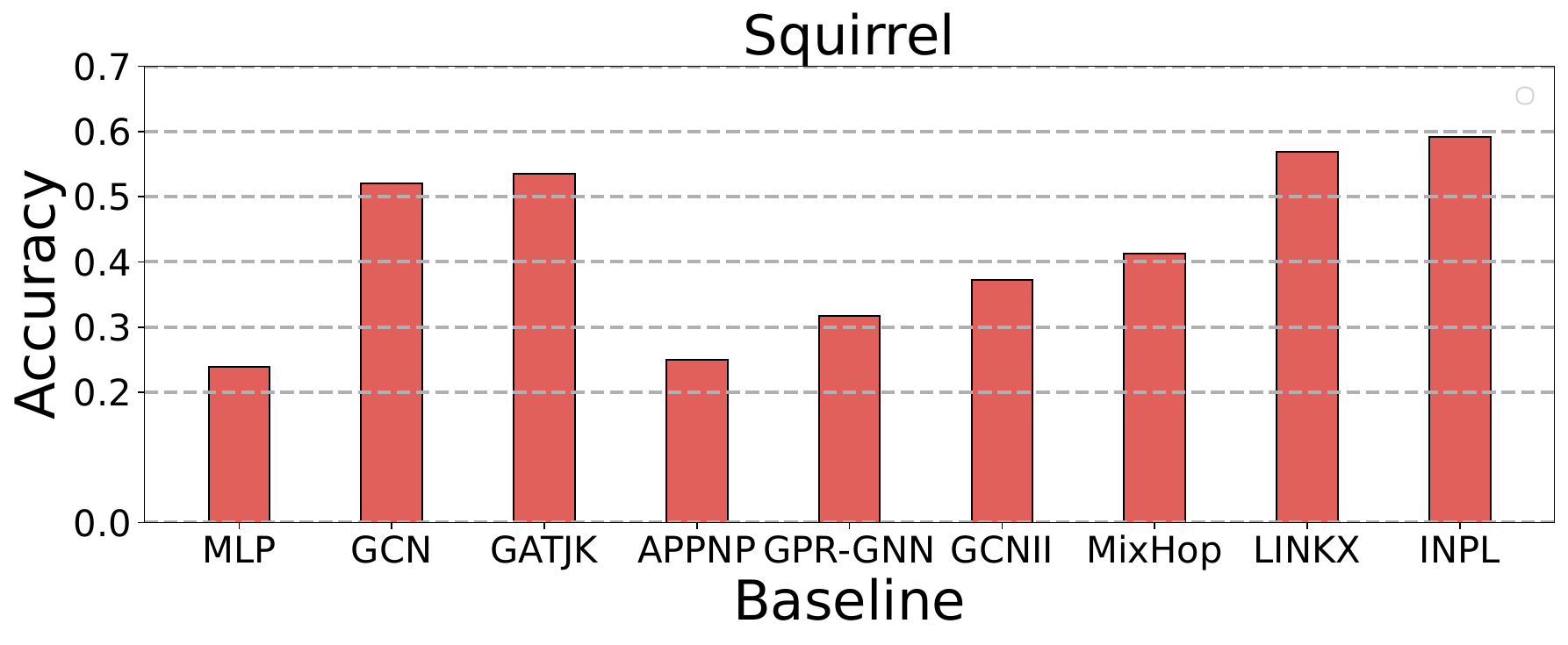}  
\end{subfigure}
\caption{Results of INPL and different baselines under neighborhood pattern-related high-bias environments for the task of semi-supervised node classification on squirrel. Compared with different baselines, our method INPL improves the accuracy of node classification under neighborhood pattern-related high-bias environments.}    
\label{figneighborhoodhighbias2}       
\end{figure*}

\subsection{Performance in unknown environments}
\label{sec:unknown _environments}
We provide the detailed results of INPL and LINKX in different unknown environments.
Note that the main paper presents the results in low-bias and high-bias environments in Figure 10, the specific statistics and results of the low-bias and high-bias environments are in Table \ref{tab:bias statistics}
and Table \ref{tab:datasets statistics1} - \ref{tab:datasets statistics4} show the statistics and results in different unknown environments ranging from low-bias to high-bias.

Here we conduct more experiments on high-bias environments to demonstrate the framework exactly alleviates the distribution shifts problem, including degree-related high-bias environments and neighborhood pattern-related high-bias environments.
\begin{itemize}
    \item \textbf{Experiments on degree-related high-bias environments}. To construct degree-related high-bias environments, we keep the nodes in the testing set unchanged and only select nodes in a specific small range of degrees as the training set, so there are strong degree-related distribution shifts between training and testing. Our main paper has shown the results of INPL and LINKX on Penn94, film, squirrel and chameleon in Figure 10, here we show the results of more baselines and additional datasets of twitch-gamers and pokec in Table \ref{tab:highbias-baselineresults}.
    \item \textbf{Experiments on neighborhood pattern-related high-bias environments.} To construct neighborhood pattern-related high-bias environments, we keep the nodes in the testing set unchanged and only select nodes whose node homophily (What percentage of neighbor nodes are at the same classes) is less than 0.5 (we call heterophilic node) as the training set, so there are strong neighborhood pattern-related distribution shifts between training and testing. The results on chameleon, pokec and squirrel are shown in Figure \ref{figneighborhoodhighbias} and Figure \ref{figneighborhoodhighbias2}.
    
\end{itemize}

From the results, we find that INPL outperforms different baselines under degree-related high-bias environments and neighborhood pattern-related high-bias environments. Although the performance of INPL also degrades under high-bias environments compared to the low-bias environments (the results of low-bias environments are shown in Table
1 in the main paper), 
the degradation is smaller compared to the other baselines for all datasets. This demonstrates the proposed INPL framework exactly alleviates the distribution shifts problem in different biased environments and learns invariant graph representation.

\subsection{Performance on homophilous datasets}
We conduct experiments on both non-homophilous datasets and homophilous datasets. Here we report our proposed INPL on the homophilous datasets of Cora, Citeseer and Pubmed \cite{yang2016revisiting} in Table \ref{tab:homophilous_results}, with the standard 48/32/20 training, validation, and test proportions. We find INPL not only outperforms LINKX but also has comparable performance to homophilous methods, such as GCN and GAT. This demonstrates that our method is also effective on homophilous datasets.

\begin{table*}[ht]
    \centering
	\caption{Comparison of homophilous datasets. Results other than LINKX reported from \cite{zhu2020beyond}. The best result per dataset is highlighted. } 
\label{tab:homophilous_results}
    {
    \begin{tabular}{ccccc}
    \toprule
	 & CiteSeer & PubMed & Cora\\
	 \# Nodes & 3327 & 19717 & 2708\\
    \midrule
	 GCN &\bestcell  $76.68 \pm{1.64} $ & $87.38 \pm{0.66} $ &  $87.28 \pm{1.26}$ \\ 
	 GAT &  $75.46 \pm{1.72}$ & $84.68 \pm{0.44} $ & $82.68 \pm{1.80} $\\ 
	LINKX & $73.19 \pm{0.99}$ &  $87.86 \pm{0.77} $ & $84.64 \pm{1.13}$  \\
     \midrule
   INPL & $76.16 \pm{0.58}$ &\bestcell  $87.97 \pm{0.12} $ &\bestcell $87.55 \pm{0.46}$  \\
	 \bottomrule
    \end{tabular}
    }
\end{table*}

\subsection{Ablation study in high-bias environments}
\label{sec:Ablation_study2}

We conduct an ablation study by changing parts of the entire framework. Specifically, we compare INPL with the following three variants:
\begin{itemize}
    \item \textbf{INPLwoLayer}: INPL without Adaptive Neighborhood Propagation Layer module.
    \item \textbf{INPLwoVar}: INPL without the variance penalty.
    \item \textbf{INPLwoKmeans}: using random graph partition instead of environment clustering.
\end{itemize}

\begin{figure*}[t]
\centering
\begin{subfigure}{0.24\linewidth}
     \includegraphics[scale=0.23]{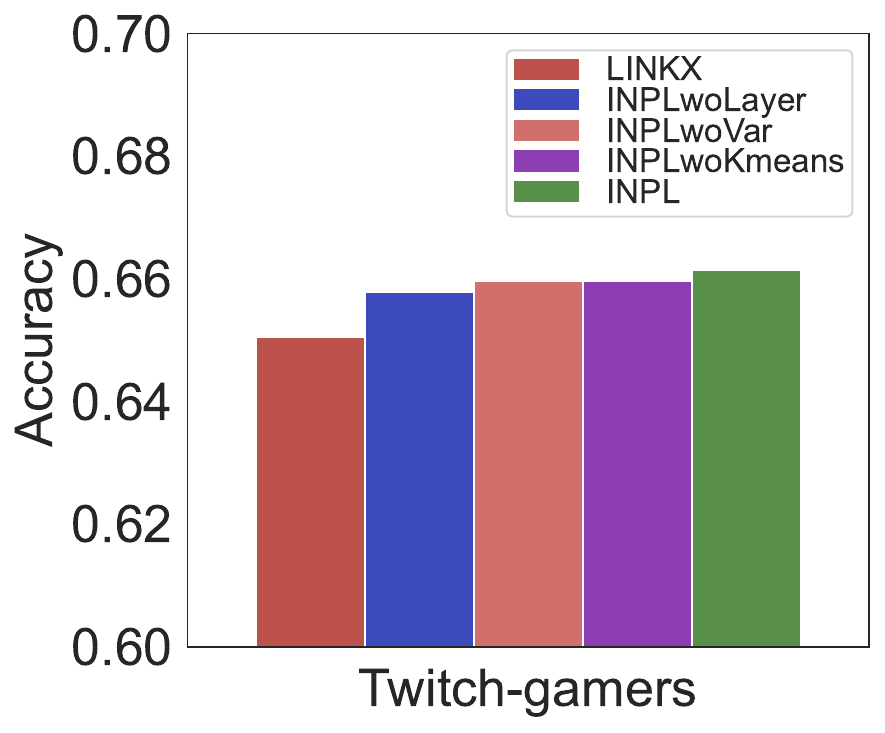}  
\end{subfigure}
\begin{subfigure}{0.24\linewidth}
     \includegraphics[scale=0.23]{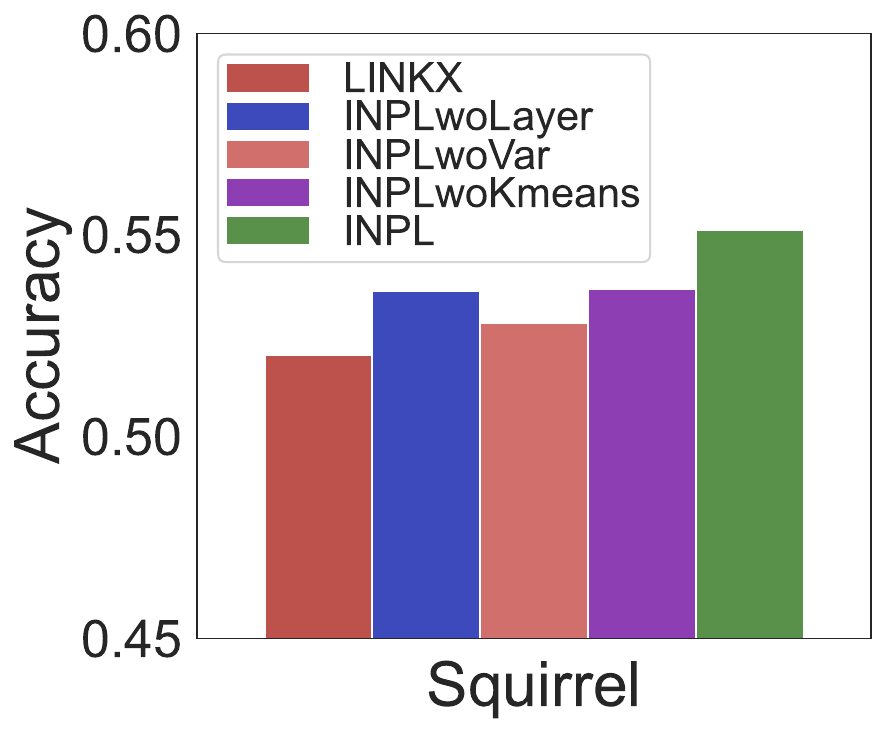}  
\end{subfigure}
\begin{subfigure}{0.24\linewidth}
     \includegraphics[scale=0.23]{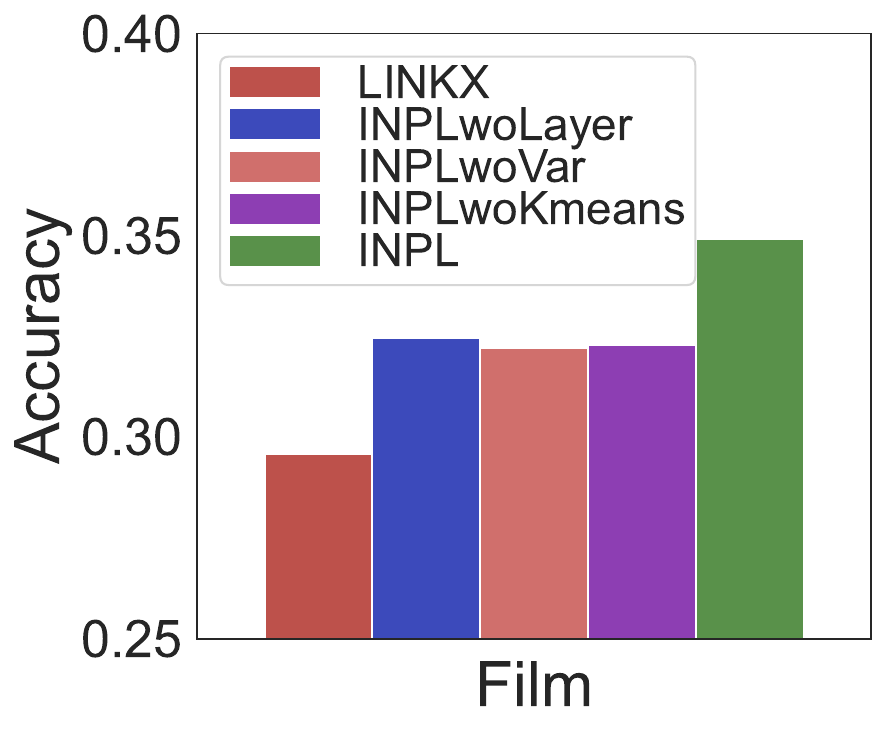}  
\end{subfigure}
\begin{subfigure}{0.24\linewidth}
     \includegraphics[scale=0.23]{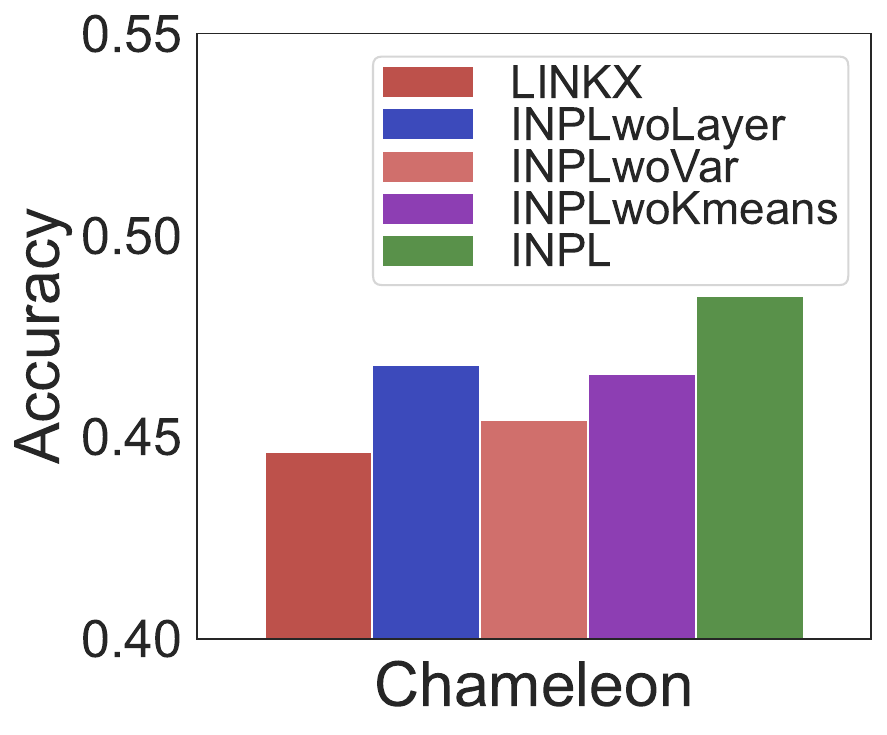}  
\end{subfigure}
\caption{Ablation study on Twitch-gamers, Film, Squirrel and Chameleon under degree-related high-bias environments. There is a drop when we remove any module of INPL, but INPLwoLayer, INPLwoVar, and INPLwoKmeans still outperform LINKX.}    
\label{highbiasfigablation}       
\end{figure*}

Note that the main paper has shown the overall performance of our INPL framework on Penn94, Film, Squirrel and Chameleon in Figure 11, which is trained in a low-bias environment. Here we conduct additional experimental results under degree-related high-bias environments, the statistics of degree-related high-bias environments are in Table \ref{tab:highbias-baselineresults}, the results are shown in Figure \ref{highbiasfigablation}. 

From the experimental results on high-bias environments, we could find that there is a drop when we remove any proposed module of INPL, but INPLwoLayer, INPLwoVar, and INPLwoKmeans still outperform LINKX, which demonstrates the effectiveness of our proposed all modules.

\section{Further discussion and limitations}
\label{discusstion}
In this paper, we conduct a comprehensive and empirical study of a novel problem, which focuses on neighborhood pattern distribution shifts problem, to the best of our knowledge, there are no studies that analyze the problem of distribution shifts of neighborhood patterns. To alleviate the neighborhood pattern distribution shifts problem on non-homophilous graphs, we propose the Adaptive Neighborhood Propagation method, where the Invariant Propagation layer is proposed to combine both the high-order and low-order neighborhood information. And adaptive propagation is proposed to capture the adaptive neighborhood information. However, we are still facing some challenges: 1) it is difficult to directly calculate higher-order matrices $A_k$ for larger datasets, since the model may not fit in GPU memory; 
2) the proposed components of INPL probably show comparable advantages on graph data, and may be less applicable or perform worse on image or text data; 3) hyper-parameters may influence the final outcome depending on possible trade-offs resulting from experiments. 
 Furthermore, we hope that the empirical observations made in this work will inspire more investigation in this direction. 

\section{Broader Impact}
\label{broader_impact}
This work has the following potential positive impact on society. Firstly, it could improve our understanding of distribution shifts on non-homophilous graphs and open up new directions for developing robust graph neural networks that are resilient to distributional changes. Secondly, these advancements could be applied in fields like social network analysis, recommendation systems, natural language processing, and image recognition, among others. 
At the same time, this work may have some negative consequences because unethical actors might use machine learning techniques developed in the above applications to spread misinformation, deceive users, amplify prejudices, personal data, or manipulate public opinion. To mitigate these risks, transparent and interpretable machine learning models should be adopted wherever possible, and developers must adopt strict guidelines for handling user data responsibly while ensuring privacy, security, and fairness in all their algorithms and products. 





\begin{align}
h^{(1)}_{v} = COMBINE\left( AGGR\{h^{(0)}_{v}\}; AGGR\{h^{(A_1)}_{u}, u \in \bar{N}_{1}(v)\};AGGR\{h^{(A_2)}_{u}, u \in \bar{N}_{2}(v)\}, ...\right) \\
AGGR\{X\} = WX + X
\end{align}
